%% file: main.tex
\documentclass[lettersize,journal]{IEEEtran}
\usepackage{amsmath,amsfonts,amssymb}
\usepackage{algorithmic}
\usepackage{algorithm}
\usepackage{array}
\usepackage{textcomp}
\usepackage{stfloats}
\usepackage{url}
\usepackage{verbatim}
\usepackage{graphicx}
\usepackage{bm}
\usepackage{cite}
\usepackage{booktabs}
\usepackage{tablefootnote}
\usepackage{multirow}
\usepackage{colortbl}
\usepackage{diagbox}
\usepackage{threeparttable}
\usepackage{mathtools}
\usepackage{subfigure}
\usepackage{xcolor}
\definecolor{linkblue}{rgb}{0, 0.19, 0.32}
\usepackage[colorlinks, linkcolor=black]{hyperref}
\hypersetup{
    colorlinks=true, %设置链接的颜色
    urlcolor=linkblue
    }

\newcommand \transpose {\mathsf{T}}

\definecolor{myblue}{RGB}{0, 0, 255}

\begin{document}

\input{contents/main_text}
\input{contents/appendix}

\end{document}

%% file: contents/main_text.tex
\title{Robotic In-Hand Manipulation for \\ Large-Range Precise Object Movement: \\ The RGMC Champion Solution}

\author{Mingrui Yu,
        Yongpeng Jiang,
        Chen Chen,
        Yongyi Jia,
        and Xiang Li,~\IEEEmembership{Senior Member,~IEEE}
        % IEEE Publication Technology,~\IEEEmembership{Staff,~IEEE,}
        % <-this % stops a space
% \thanks{Manuscript received April 19, 2021; revised August 16, 2021.}
\thanks{This work was supported in part by the Science and Technology Innovation 2030-Key Project under Grant 2021ZD0201404, in part by the National Natural Science Foundation of China under Grant 623B2059 and U21A20517, in part by the BNRist project under Grant BNR2024TD03003, and in part by the Institute for Guo Qiang, Tsinghua University.}
\thanks{M. Yu, Y. Jiang, C. Chen, Y. Jia, and X. Li are with the Department of Automation, Tsinghua University, Beijing, China. Corresponding author: Xiang Li (xiangli@tsinghua.edu.cn).}
}

% The paper headers
\markboth{Journal of \LaTeX\ Class Files}%
{Shell \MakeLowercase{\textit{et al.}}: A Sample Article Using IEEEtran.cls for IEEE Journals}

\IEEEpubid{0000--0000/00\$00.00~\copyright~2021 IEEE}
% Remember, if you use this you must call \IEEEpubidadjcol in the second
% column for its text to clear the IEEEpubid mark.

\maketitle

\begin{abstract}
In-hand manipulation using multiple dexterous fingers is a critical robotic skill that can reduce the reliance on large arm motions, thereby saving space and energy. This letter focuses on in-grasp object movement, which refers to manipulating an object to a desired pose through only finger motions within a stable grasp. The key challenge lies in simultaneously achieving high precision and large-range movements while maintaining a constant stable grasp. To address this problem, we propose a simple and practical approach based on kinematic trajectory optimization with no need for pretraining or object geometries, which can be easily applied to novel objects in real-world scenarios. Adopting this approach, we won the championship for the in-hand manipulation track at the 9th Robotic Grasping and Manipulation Competition (RGMC) held at ICRA 2024. Implementation details, discussion, and further quantitative experimental results are presented in this letter, which aims to comprehensively evaluate our approach and share our key takeaways from the competition. Supplementary materials including video and code are available at \texttt{\url{https://rgmc-xl-team.github.io/ingrasp_manipulation}}.
\end{abstract}


\section{Introduction}

\IEEEPARstart{I}{n-hand} manipulation with multi-fingered hands has become increasingly important in recent research on robotic manipulation, as it is crucial for achieving human-level dexterity \cite{billard2019trends}. Although the advantages of utilizing the high degrees of freedom (DoFs) of multi-fingered hands are attractive, coordinating the fingers to efficiently and robustly manipulate in-hand objects as expected in real-world environments remains a challenging and unresolved issue.

\begin{figure} [tb]
  \centering 
    \includegraphics[width=1.0\linewidth]{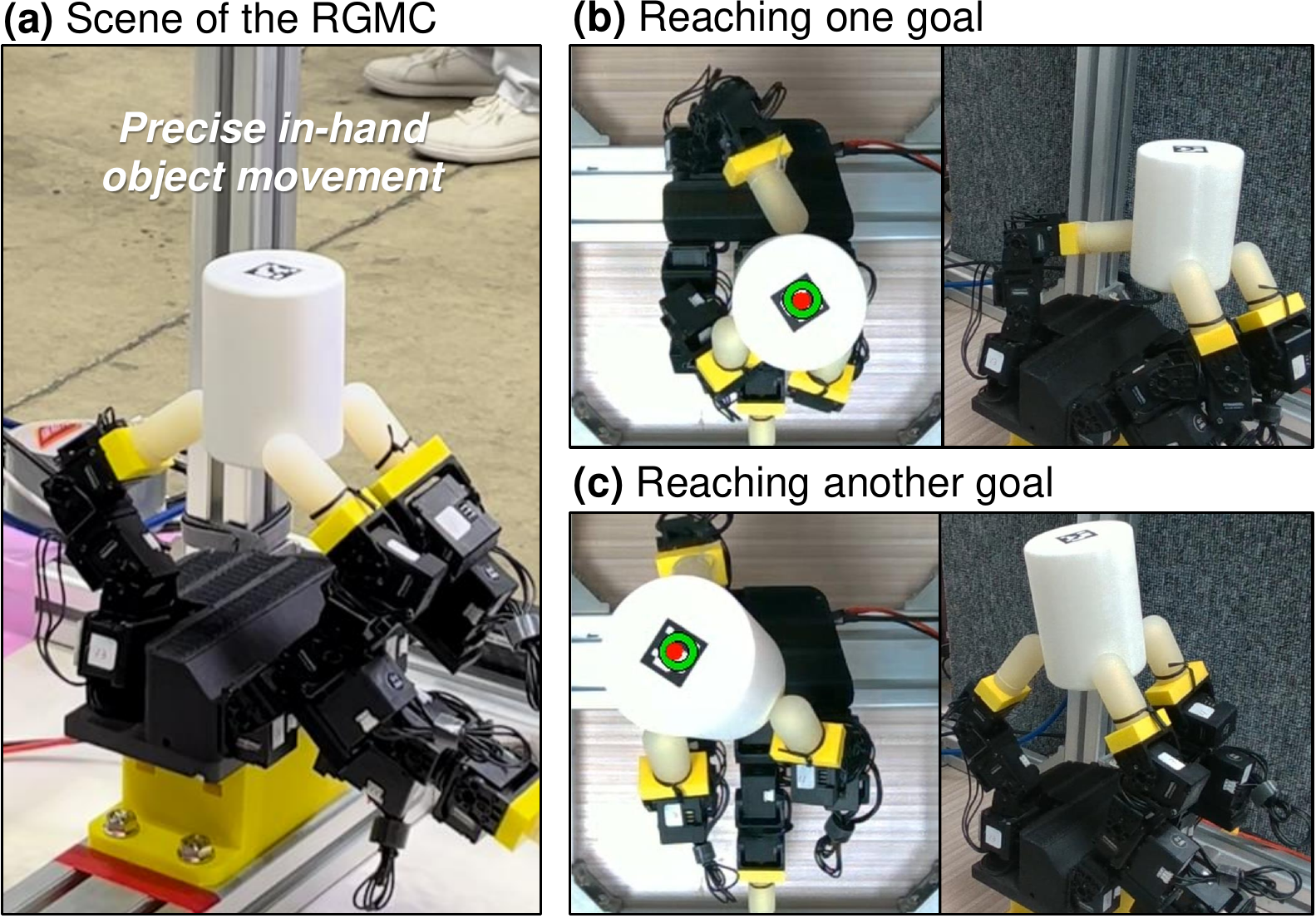} 
  % \vspace{-2mm}
  \caption{In-grasp object movement task, where the goal is to manipulate the in-hand object to a desired pose (position) using only finger motions within a stable grasp. 
  (a) Scene of the Robotic Grasping and Manipulation Competition (RGMC) at ICRA 2024, where we won the championship for this task.
  (b)(c) Precisely moving the object to the desired position in a large in-hand space. 
  }
  \label{fig:figure1}
    \vspace{-3mm}
\end{figure}

% \cite{yu2024hand,zhou2024hand,andrychowicz2020learning}
Amid a diverse range of in-hand manipulation tasks, this letter focuses on a fundamental task, namely \textit{in-grasp object movement}, which refers to controlling an object's pose (position or orientation) relative to the hand using only finger motions within a stable grasp \cite{hang2020hand,hang2021manipulation,grace2024direct,sundaralingam2017relaxed,sundaralingam2019relaxed}, as shown in Fig. \ref{fig:figure1}.
This task was benchmarked in the in-hand manipulation track of the 9th Robotic Grasping and Manipulation Competition (RGMC) held at ICRA 2024 \cite{rgmc2024}.
Although robot arm motions alone can sometimes achieve similar desired object movements, they consume much more energy than finger motions. Moreover, achieving small desired object movements may require large arm joint motions, which can be problematic in constrained spaces with obstacles.

% necessary for IEEE copyright
\IEEEpubidadjcol

The significance of \textit{in-grasp} manipulation is that, for many tasks, desired in-hand object motions can be achieved without altering the contact locations. Compared with general in-hand manipulation involving making and breaking contacts, explicitly imposing constant stable grasp constraints can significantly reduce the complexity of problem solving. 
Moreover, avoiding contact switching can enhance robustness in real-world scenarios, as the complex mechanics of real-world dynamic contacts may exacerbate the sim-to-real (model-to-real) gap. 

The challenges in this task stem from several factors.
First, a stable grasp with constant contact locations must be strictly maintained throughout the manipulation process; otherwise, the object will fall.
Second, the reachable space of the in-grasp object is highly constrained by the limited DoFs (usually $\leq 4$) and the joint limit of each finger, especially when the stable grasp constraint is maintained between fingers.
Third, real-world task accuracy is affected by imperfect hand kinematics, imperfect hand control, and modeling gap of contacts.
Consequently, achieving precise and large-range in-grasp object movement in the real world is non-trivial.

To address the above challenges and achieve deployable multi-fingered in-grasp object movement, this letter proposes a simple and practical approach based on trajectory optimization.
Specifically, to ensure a constant stable grasp during manipulation and generalization across different goals and objects, we employ geometry-free trajectory optimization with explicit constraints rather than reinforcement learning, which may lack control over accuracy. 
To expand the object's reachable space, we use global full-trajectory planning and allow rolling contact between the object and all fingertips. 
Furthermore, to compensate for simplifications in the trajectory optimization and enhance real-world accuracy, we incorporate a closed-loop scheme via re-planning and re-execution. 

Our key contributions are highlighted as follows:
\begin{enumerate}
    \item We propose a simple and practical trajectory optimization approach for multi-fingered in-grasp object movement. 
    % The approach is formulated concisely and easy to implement. 
    The approach does not rely on large-scale training, complex contact models, or even object geometries, making it easily applicable to real-world novel objects. Compared with existing works, it achieves a larger object reachable space while ensuring task accuracy.

    \item Beyond the validation of our solution through the RGMC, this letter presents a detailed quantitative analysis of our approach through extensive real-world experiments. Moreover, we share the implementation details (including the source code) and practical insights gained from the competition.
\end{enumerate}

Thanks to its accuracy, robustness, and generalizability, our approach won the championship of the in-hand manipulation track of the RGMC \cite{rgmc2024}. Additionally, our approach was awarded the Most Elegant Solution among all tracks of the RGMC, owing to its concise and novel formulation.

\section{Related Work}

In this section, we introduce existing approaches to in-hand manipulation and provide a comparison with our proposed approach to in-grasp precise manipulation. 

\subsection{General Multi-Fingered In-Hand Manipulation}

Without explicit constraints of contacts, the general in-hand manipulation tries to manipulate the object through any possible finger motions and contact sequences.

Recently, learning-based approaches have been successfully applied to dexterous manipulation. 
Reinforcement learning (RL) has shown excellent performance for in-hand manipulation that requires complex dynamic finger gaiting, with the in-hand object rotation being a representative task \cite{andrychowicz2020learning,qi2023general,chen2023visual,pitz2024learning}. By training on large-scale interactions, RL-based approaches can learn complex multi-fingered behaviors that are difficult to model and plan. 
However, RL can be costly for in-grasp manipulation. First, it must rely on inefficient random exploration during the early training phase to learn how to maintain a constant stable grasp \cite{hu2023dexterous}, which is a prerequisite for subsequent object movement. 
Second, generalization to arbitrary goals and objects is expensive to guarantee, as it can only be improved by appreciably expanding the training dataset to implicitly cover any possible test case \cite{chen2023visual}.
Third, unlike in-hand rotation that allows for larger tolerance in acceptable finger motions, in-grasp object movement requires high precision for arbitrary goal poses, which is an objective at which RL does not excel.

Human skill and experience can also be leveraged in dexterous manipulation, e.g., through imitation learning (IL) \cite{wang2024dexcap,ze20243d,guzey2024see}.
Although IL performs well on daily tasks whose goals are hard to mathematically define, it is less effective for precise in-grasp object movement, as precisely controlling the object by teleoperation is difficult for humans lacking complete in-hand feedback. The generalization is also difficult to guarantee.

In parallel, considerable works have been done on model-based in-hand manipulation, which can be divided into two categories.
The first is contact-explicit approaches, which first explicitly search for discrete contact sequences and then control the hand to track the planned contacts \cite{chen2021trajectotree,zhu2023efficient,cheng2024enhancing}. Our approach for in-grasp manipulation falls into this category, with contacts being predefined by the initial grasp.
The second category is contact-implicit approaches, which use unified contact-motion models with contact smoothing to efficiently explore potential finger motions and resulting contacts \cite{pang2023global,jiang2024contact,jin2024complementarity}.
These approaches excel at efficiently determining contact sequences but sacrifice physical fidelity, making them less suitable for precise in-grasp manipulation.

\subsection{In-Grasp Manipulation}
As a typical type of in-hand manipulation, in-grasp manipulation refers to controlling the object’s pose within a stable grasp.
One of the key challenges is to strictly maintain a constant stable grasp while moving the object.
A series of representative works \cite{hang2020hand,hang2021manipulation,grace2024direct} on in-grasp object movement used self-designed under-actuated fingers with compliant passive joints (enabled by springs), which could naturally adapt to the in-hand object and ensure a stable grasp during random finger motion. Although such end-effectors are effective for in-grasp object movement, they may lack versatility for other types of in-hand manipulation.
In contrast, we use an open-sourced, full-actuated anthropomorphic hand, the Leap Hand \cite{shaw2023leaphand}, which is a low-cost generic end-effector widely used for various tasks. 
Consequently, explicitly constraining the computed joint motion to maintain a constant stable grasp is essential. 
Additionally, they used a local mapping from robot motion to object motion to iteratively reach local goals (usually $\leq 2$ cm). In contrast, we optimize the full trajectory to arbitrary large-range goals within a $5 \times 5 \times 5$ (cm) cubic space, as required by the RGMC.

Classically, the constraint of constant stable grasping is usually analyzed at the velocity level. Some works used the velocity constraint to rigorously convert the desired object velocity into finger joint velocities \cite{hartl1995dexterous} or convert the desired object velocity and acceleration into finger joint torques \cite{li1989grasping}. 
These approaches were validated in simulation and not easy to deploy in the real world, since they required fully predefined object trajectories and accurate system information, including contact surface parameters or hand-object dynamics.

Our approach was initially inspired by \cite{sundaralingam2017relaxed,sundaralingam2019relaxed} that were based on purely kinematic trajectory optimization with pose constraints. To generate in-grasp trajectories, they assumed rigid contact between the thumb-tip and object (i.e., an invariant relative pose) and imposed relaxed-rigidity constraints on the other fingertips (i.e., penalizing changes in fingertip poses in the thumb-tip frame). However, the assumption of rigid thumb-tip contact is overly restrictive, as it completely forbids rolling between the thumb-tip and object. This severely limits the object's moving space owing to the low DoFs and joint limit of the thumb.
In contrast, we allow rolling contacts of all fingertips by fixing only the fingertip positions in the object frame, which enlarges the reachable space of the object.

Although we allow fingertip rolling, we do not include the rigorous rolling constraint in the optimization like a recent work \cite{yang2024multi} that required the precise geometry of the object and fingertips.
Moreover, the velocity-level geometry-aware rolling constraint greatly increases the complexity of the optimization, making it impractical for real-time deployment (e.g., requiring around 8 seconds per model predictive control step in \cite{yang2024multi}).

\section{Preliminaries}

\subsection{Competition Task Setup}

The competition task involves two objects: a known cylindrical object and a novel object.
The poses of the object are tracked using an attached AprilTag marker. The initial grasp is set by a human operator with no restrictions. 
For the known object, cylinders of three sizes are provided, and each team can select one based on their hand design. Using the hardware shown in Fig. \ref{fig:hardware}, we chose the smallest cylinder with a diameter of 60 mm and a height of 80 mm. 
The novel object is randomly assigned on-site during the competition.
Goal waypoints are given as a sequence of 10 positions relative to the initial object's (AprilTag's) position. After forming the initial grasp, the hand is tasked to autonomously and continuously move the object (AprilTag) through the waypoints one by one. These waypoints are sampled within a $5 \times 5 \times 5$ (cm) cubic space centered at the initial object position. Goal object orientations are not assigned in the competition. 
Each goal waypoint should be reached within a 20-second time budget. The evaluation metric is the accumulated position error across the 10 waypoints. 
If the object is dropped, it cannot be manually reset to continue with the remaining waypoints.
The best of two runs on each object is used to rank the teams. More details of the competition rules can be found in \cite{inhand2024rgmc}.

\subsection{Notations and Definitions} \label{section:notations}

In this letter, $\bm q_{i, t}$ represents the joint position vector of the $i^{\rm th}$ finger at time $t$. For convenience, we denote $\bm Q_{t} = [\bm q_{1, t}; \cdots; \bm q_{n, t}]$, where $n$ is the number of fingers in grasp. 
The pose of the $i^{\rm th}$ fingertip at time $t$ is denoted as $\bm T_{i, t}$.
The object pose at time $t$ is denoted as $\bm T_{\text{o}, t}$, with the position and orientation respectively denoted as $\bm p_{\text{o}, t}$ and $\bm R_{\text{o}, t}$.
The desired object pose is denoted as $\bm T_{\rm o, d}$.
% , with the desired position and orientation respectively denoted as $\bm p_{\rm o, d}$ and $\bm R_{\rm o, d}$ (optional).
The world frame (coordinates) and object frame are denoted as $\mathcal{W}$ and $\mathcal{O}$, respectively. 
The notations are summarized in Fig. \ref{fig:sketch}.
We use $[\bm a; \bm b]$ to denote the vertical concatenation of column vectors $\bm a$ and $\bm b$.
Following the theory of Lie groups and Lie algebra for robotics \cite{murray1994mathematical}, we denote the conversion from axis-angle rotation vector $\bm r \in \mathfrak{so}(3)$ to rotation matrix $\bm R \in \text{SO}(3)$ as $\bm R = \exp (\bm r^{\wedge})$ and the inverse conversion as $\bm r = \ln (\bm R)^{\vee}$.

In this letter, we define the weighted scalar distance $d$ between two poses $\bm T_1$ and $\bm T_2$ as
\begin{equation} \label{eq:distance_definition}
    d(\bm T_1, \bm T_2, \bm W) = 
    \frac{1}{2} \bm e^\transpose \bm W \bm e
    = \frac{1}{2} \bm p_{\rm e}^\transpose \bm W_{\rm p} \bm p_{\rm e} + \frac{1}{2}\bm r_{\rm e}^\transpose \bm W_{\rm r} \bm r_{\rm e},
\end{equation}
where $\bm e = [\bm p_{\rm e}; \bm r_{\rm e}] \in \mathfrak{se}(3)$, in which $\bm p_{\rm e} = \bm p_1 - \bm p_2$ is the position error vector, 
and $\bm r_{\rm e} = \ln \left( \exp(\bm r_1^{\wedge}) \left(\exp (\bm r_{2}^{\wedge}) \right)^{-1} \right)^{\vee}$ is the rotation error vector.
In addition, $\bm W = \text{diag}(\bm W_{\rm p}, \bm W_{\rm r})$ is a weighting matrix for different dimensions of the errors.

\section{Method}

\subsection{Hardware Setup}

\begin{figure} [tb]
  \centering 
    \includegraphics[width=0.95\linewidth]{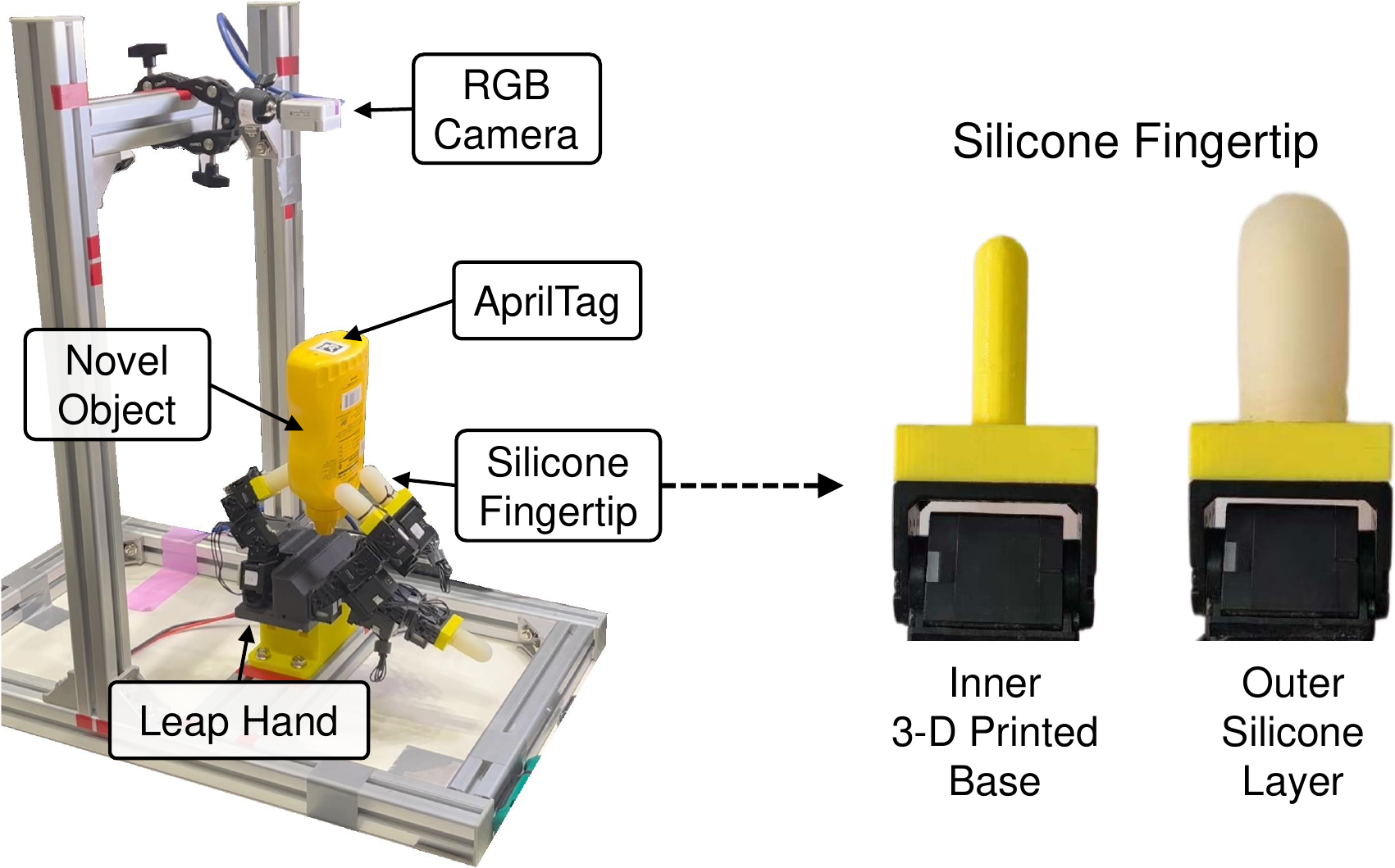} 
  % \vspace{-2mm}
  \caption{Our hardware setup for the competition, comprising a Leap Hand with soft silicone fingertips, a top-view RGB camera, and an in-hand object with an AprilTag marker. Each silicone fingertip comprises an inner 3-D printed base and an outer silicone layer.}
  \label{fig:hardware}
    \vspace{-3mm}
\end{figure}

The hardware setup is shown in Fig. \ref{fig:hardware}. 
A calibrated top-view RGB camera (RealSense d405) is used to obtain the AprilTag pose.
We use a Leap Hand \cite{shaw2023leaphand}, which features four fingers, each with four actuated DoFs.
The control commands to the hand are the target joint angles, which are tracked by a low-level PD controller.
We only use the thumb, index, and ring fingers for this task, as three fingers with soft frictional contacts are sufficient for a stable grasp. Including additional fingers increases the risk of self-collision and reduces the overall finger workspace. 
To enhance contact compliance and friction, we replace the original fingertips with custom-made soft silicone fingertips.
The contacts between these hemispherical fingertips and the object approximate the point contacts with friction. 
From a hardware perspective, both the low-level PD controller and the soft fingertips provide physical compliance, facilitating stable grasping during manipulation.

\subsection{Trajectory Optimization}

\begin{figure} [tb]
  \centering 
    \includegraphics[width=0.9\linewidth]{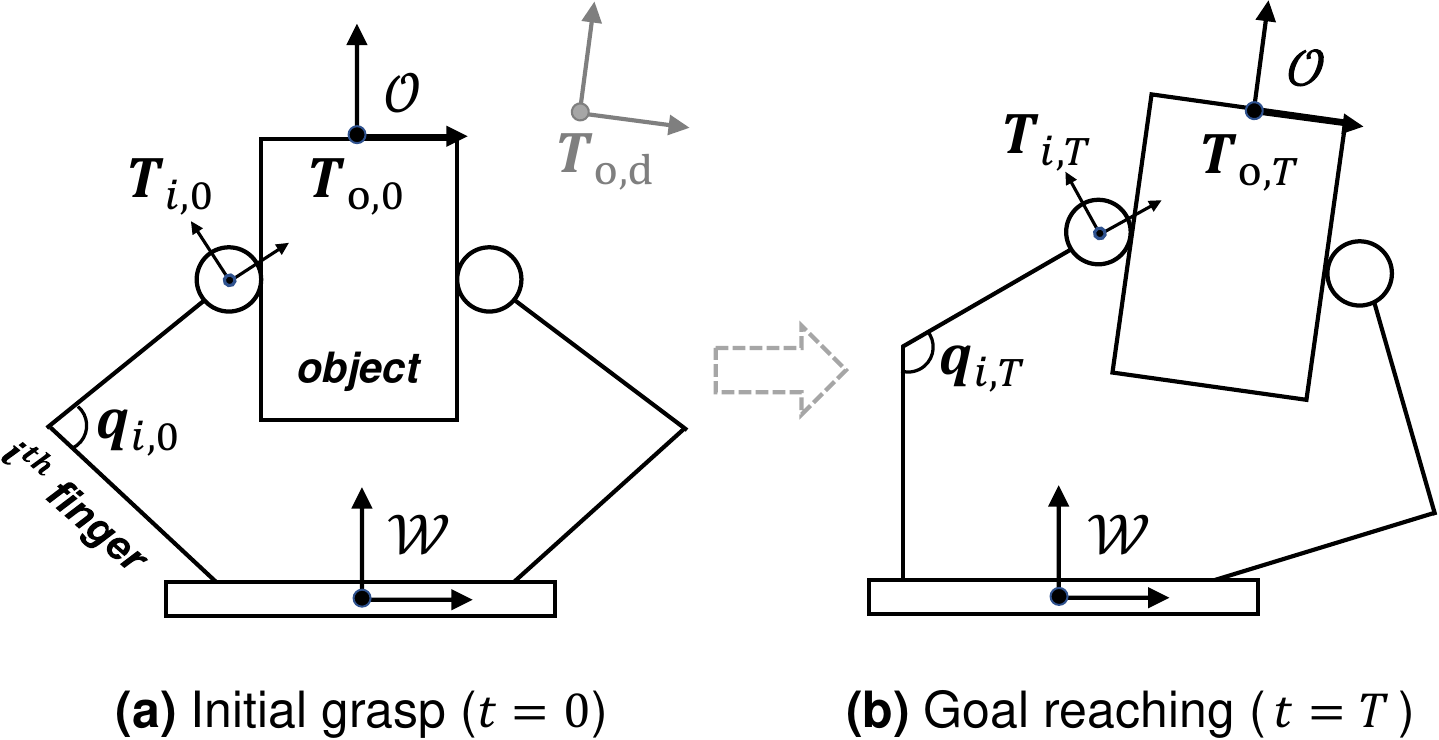} 
  \vspace{-2mm}
  \caption{Formulation of the in-grasp object movement. The objective is to find a hand joint trajectory to move the object, starting from $\bm T_{\text{o}, 0}$ and reaching $\bm T_{\text{o}, \text{d}}$ at time step $T$ while maintaining a constant stable grasp.
  }
  \label{fig:sketch}
  \vspace{-3mm}
\end{figure}

We define the full trajectory as a sequence of $T+1$ points.
As illustrated in Fig. \ref{fig:sketch}, the objective of the trajectory optimization is to find a hand joint trajectory $\bm Q_{1:T}$ to move the object from the initial configuration ($\bm Q_{0}$ and $\bm T_{\text{o}, 0}$) to the desired object pose $\bm T_{\rm o, d}$ as closely as possible at the end of the trajectory, during which the fingers maintain a stable grasp and avoid self-collisions. 

We consider this problem under the following assumptions:
\begin{enumerate}
    \item The in-hand object is rigid.
    \item The initial grasp (defined by $\bm Q_{0}$ and $\bm T_{\text{o}, 0}$) is a stable and manipulable grasp.
    \item The object surface near the grasp contact points is smooth and exhibits low curvature.
    \item The hand and object move at a slow speed, allowing the manipulation process to be treated as quasi-static with negligible inertial effects.
\end{enumerate}

The core idea of our trajectory optimization approach is concise: trying to minimize the terminal object pose error while maintaining constant contact locations. 
However, constraining the exact contact locations is challenging, as it requires precise initial contact locations and high-fidelity geometries of the object and fingertips, which are hard to obtain in practice.
Consequently, we simplify the constant stable grasp constraint as fixing the fingertip positions (i.e., the center of the hemispherical tips) in the object frame $\mathcal{O}$. 
Note that this simplification implies that we ignore the small changes in contact positions resulting from rolling between the object and fingertips. 
We assume that, for objects with low-curvature surfaces, minor changes in contact positions caused by rolling will not significantly affect the manipulation result.

The trajectory optimization problem is specifically formulated as follows.
The termination cost regarding the desired object pose is defined as
\begin{equation}
\mathcal{J}_{\rm object} = d({^{\mathcal{W}} \bm T_{\text{o}, T}}, {^{\mathcal{W}} \bm T_{\rm o, d}},
\bm W_{\rm o} ) ,
\end{equation}
where the distance $d(\cdot)$ is defined in (\ref{eq:distance_definition}). If the goal object orientation is not specified, we can assign zero orientation weights to the weighting matrix $\bm W_{\rm o}$.

The cost for a constant stable grasp is defined as
\begin{equation} \label{eq:J_finger}
\mathcal{J}_{\rm finger} = \sum_{t=1}^{T} \sum_{i=1}^{n} d({^{\mathcal{O}} \bm T_{i, t}}, {^{\mathcal{O}} \bm T_{i, 0}}, \bm W_{\rm f}) ,
\end{equation}
where ${^\mathcal{O} \bm T_{i, 0}}$ is the initial fingertip pose in the object frame $\mathcal{O}$. 
Here, the constant grasp requirement is treated as a soft constraint to avoid strictly infeasible situations.
By assigning zero orientation weights to $\bm W_{\rm f}$, we can fully allow rolling.

Additionally, we include a joint-space penalty to reduce (or restrict) the joint trajectory length and make the waypoints distributed evenly: 
\begin{equation}
    \mathcal{J}_{\rm joint} = \lambda \sum_{t=0}^{T-1} \| (\bm Q_{t+1} - \bm Q_{t}) \|_2^2  , 
\end{equation}
where $\lambda$ is a scalar weight.

The trajectory optimization problem is then formulated as 
\begin{equation}
\begin{aligned}
\min_{\bm Q_{1:T}, \bm T_{\text{o}, 1:T}} & \quad \mathcal{J} = \mathcal{J}_{\rm object} + \mathcal{J}_{\rm finger} + \mathcal{J}_{\rm joint} 
\\
\text{s.t.} \quad & \bm Q^{\rm min} \preceq \bm Q_{t} \preceq \bm Q^{\rm max}, \quad \forall t \in [1, T]
\\
& \bm F_{\text{collision}}(\bm Q_{t})  \preceq \bm 0 , \quad \forall t \in [1, T] ,
\end{aligned}
\end{equation}
where the first hard constraint is the joint limit constraint, and the second avoids collisions between fingers. In particular, for the competition, we constrain the distances between four critical points selected on the index and ring fingers. 
Given the initial stable grasp, the trajectory optimization does not explicitly use the object geometry, allowing it to be applied to novel objects with no need for object reconstruction.

Note that we include the object poses $\bm T_{\text{o}, 1:T}$ in the optimization variable, different from that in \cite{sundaralingam2017relaxed} which only included the finger joint angles. This is because the object pose in their work could be represented by the thumb-tip pose under the assumption of rigid thumb-object contact, whereas our object pose cannot be inferred from finger poses due to the rolling contact. Allowing thumb-object rolling enlarges the object's reachable space.
The object orientations in the optimization variables are represented by axis-angle rotation vectors $\in \mathfrak{so}(3)$.
We solve this non-convex constrained optimization problem using the Sequential Least Squares Programming (SLSQP) \cite{kraft1988software} algorithm implemented by the \texttt{scipy.optimize.minimize} in Python. 

\subsection{Gradients of the Optimization}
Providing analytical gradients to the solver is crucial for improving solving efficiency. 
Although our optimization formulation is concise, deriving the analytical gradients requires some effort due to the inclusion of orientation variables, orientation costs, and relative poses between two moving frames.
Due to page limit, here we briefly introduce the gradient of pose distances. Details are provided in the Appendix at the \href{https://rgmc-xl-team.github.io/ingrasp_manipulation}{\textbf{project website}} to facilitate re-implementation.
Using (\ref{eq:distance_definition}), the gradient of $d(\bm T, \bm T_{\rm d}, \bm W)$ w.r.t. variable $\bm x$ is derived as
\begin{equation}
    \frac{\partial d}{\partial \bm x} 
    = \frac{\partial d}{\partial \bm e}
    \frac{\partial \bm e}{\partial \bm x} 
    = \frac{\partial d}{\partial \bm e}
    \frac{\partial \bm e}{\partial \bm \phi} 
    \frac{\partial \bm \phi}{\partial \bm x} ,
\end{equation}
where we introduce a perturbation variable $\bm \phi \in \mathfrak{se}(3)$ for convenience of calculation, which represents a left perturbation on $\bm T$. First, we have $\frac{\partial d}{\partial \bm e} = \bm e^\transpose \bm W$. Second, we can obtain 
\begin{equation}
    \frac{\partial \bm e}{\partial \bm \phi} = 
    \left[
    \begin{array}{cc}
    \bm I & \bm 0 \\
    \bm 0 & \bm J_l(\bm r_{\rm e})^{-1}
    \end{array}
    \right] ,
\end{equation}
where $\bm J_l(\cdot)$ refers to the left Jacobian matrix of SO$(3)$ \cite{barfoot2024state}.
Third, we have $\frac{\partial \bm \phi}{\partial \bm x} = \bm J(\bm x)$, where $\bm J(\bm x)$ is the space Jacobian that relates the spatial twist to $\dot{\bm x}$. 
Specifically, for $\mathcal{J}_{\rm object}$, the variable $\bm x$ is the object pose, and the space angular Jacobian can be derived as $\bm J_l(\bm r)$, where $\bm r$ is the object orientation. 
For $\mathcal{J}_{\rm finger}$, the variable $\bm x$ comprises both the object pose and finger joint angles, and the space Jacobian is the relative Jacobian between the fingertip twist in $\mathcal{O}$ and $\dot{\bm x}$ \cite{jamisola2015more}. Please refer to the \href{https://rgmc-xl-team.github.io/ingrasp_manipulation}{\textbf{Appendix}} for details.

\subsection{Closed-Loop Execution} \label{sec:closed_loop_exe}

Open-loop execution of the planned trajectory may result in low accuracy, as the optimization problem ignores small changes in contact positions due to rolling and potential slippage on the object surface. Additionally, sensing errors of the initial object pose, inaccurate finger kinematics, and joint-space control errors of the fingers, can affect actual execution results.
Consequently, we adopt a closed-loop execution scheme simply based on re-planning and re-execution.
After each iteration of trajectory optimization and actual execution, we plan and execute a new trajectory from the current state to the desired object pose. 
We repeat this re-planning and re-execution process until any of the following conditions is met: 1) the actual error is smaller than the planned error (i.e., the distance between $\bm T_{\text{o}, \text{d}}$ and planned $\bm T_{\text{o}, T}$); 2) the predefined maximum number of re-planning attempts $N_{\rm replan}$ is reached; or 3) the time budget is exceeded. This strategy helped us achieve very high precision in the competition.

\begin{figure} [tb]
  \centering 
    \includegraphics[width=1.0\linewidth]{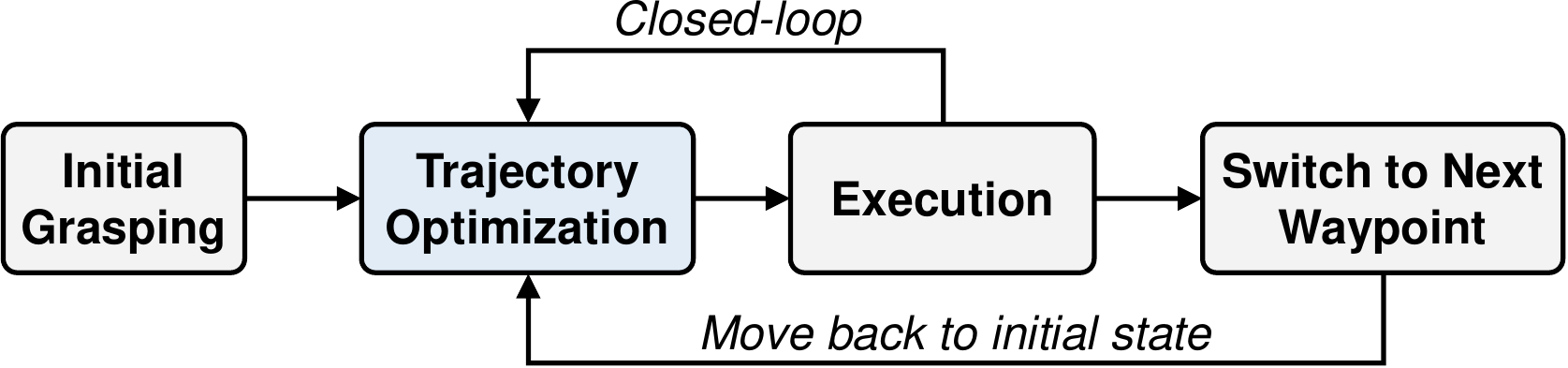} 
  % \vspace{-2mm}
  \caption{Our pipeline for the RGMC. After the initial grasping, our solution iteratively plans a path from the current state to the goal through trajectory optimization and executes it until certain conditions are satisfied. Then, it switches to reach the next waypoint, during which the hand first returns to the initial state and then moves the object to the next goal.}
  \label{fig:workflow}
    \vspace{-3mm}
\end{figure}

In addition, the competition requires continuous reaching of a series of waypoints in a large space, which necessitates long-horizon solvability and robustness. 
Our strategy is to move the fingers back to the initial state (along the forward trajectory) after reaching each waypoint, as the initial state is usually a better start for trajectory optimization to the next goal.
The overall pipeline is shown in Fig. \ref{fig:workflow}, which is used for both the competition and the experimental evaluation in Section \ref{sec:experimental_results}.

\begin{figure*} [tb]
  \centering 
    \includegraphics[width=\textwidth]{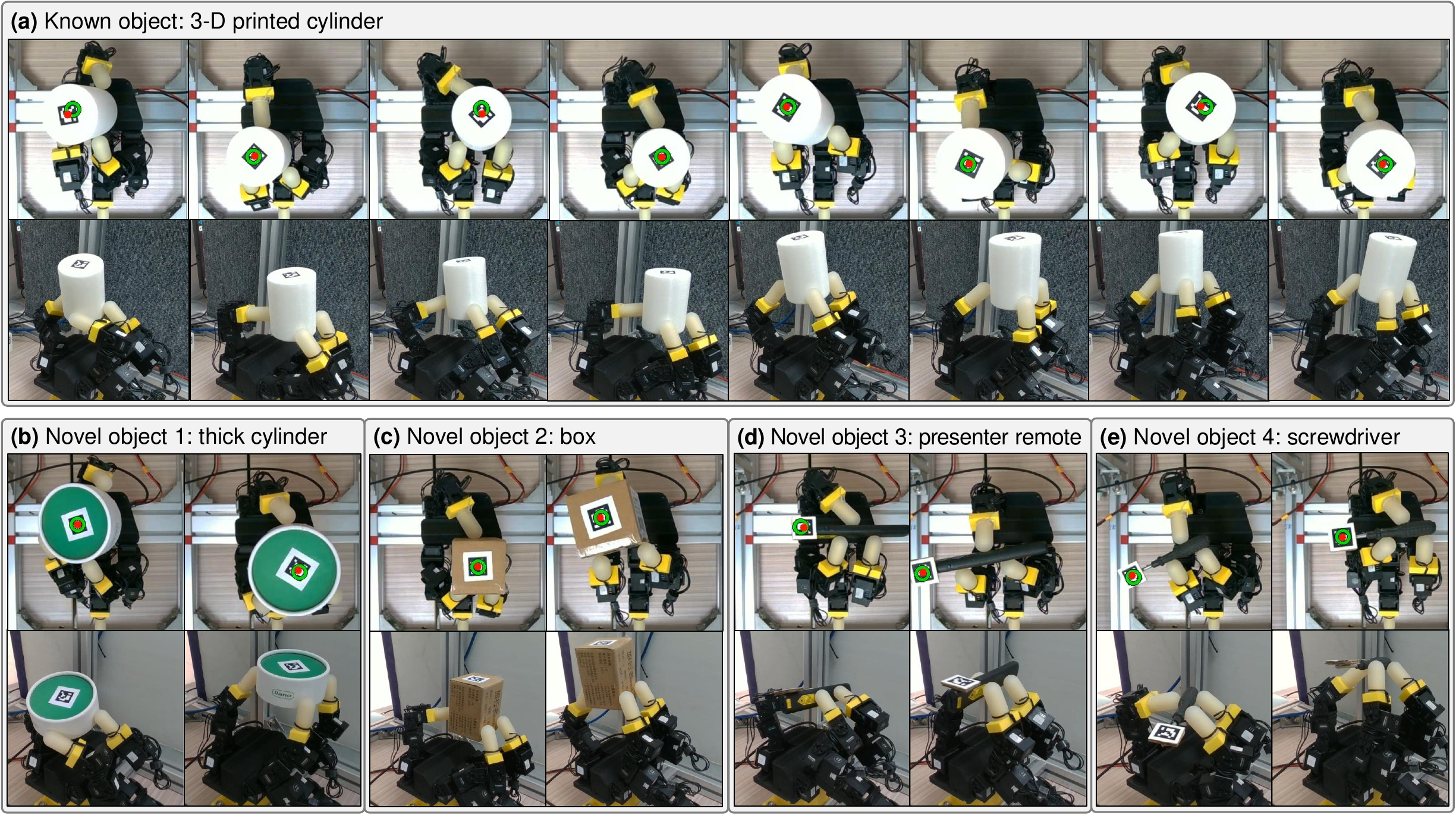} 
  \vspace{-5mm}
  \caption{Experiments of in-grasp object movement with various objects, in which the objects continuously reach the eight corners of a $5\times5\times5$ (cm) cubic space. (a) The known cylinder object provided by the competition organizer. (b) The novel everyday objects used in our experiments, including a thick cylinder lid, box, presenter remote, and screwdriver. 
  For each object, the images in the first row are from the top-view camera used for object pose tracking, where the red points and green circles represent the AprilTag centers and the desired positions, respectively; the images in the second row are from another camera used only for visualization.
  More manipulation processes are shown in the \href{https://rgmc-xl-team.github.io/ingrasp_manipulation}{\textbf{supplementary video}}.
  }
  \label{fig:objects}
    \vspace{-3mm}
\end{figure*}

\subsection{Initial Grasping} \label{sec:initial_grasp}

The quality of the initial grasp is critical for the subsequent in-grasp object movement.
For the known cylinder, we manually define the target fingertip positions for the initial grasp. 
To efficiently obtain an initial grasp of a novel object without on-site coding, we develop a human-dragging approach, in which a human can freely move the fingers along a horizontal plane, while the algorithm constrains the fingertip heights to keep them on the same plane. This enables the human to simultaneously and conveniently adjust the three fingers to establish and record a stable grasp.
Then, the target fingertip positions are set further inside the object surface with a predefined offset.
This ensures that the fingertips apply appropriate grasping forces, taking advantage of the compliance provided by the low-level PD controller and soft fingertips.
Finally, we use an optimization-based inverse kinematics (IK) solver \cite{yu2024hand} to compute the corresponding joint-space position command.

\section{Experimental Results} \label{sec:experimental_results}

First, we quantitatively analyze the performance of the proposed approach and the effects of hyper-parameters, using the known cylinder object shown in Fig. \ref{fig:objects}(a). 
Then, we validate the generalization of the approach to novel objects, using the everyday objects shown in Fig. \ref{fig:objects}(b)-(e). 
We use the following metrics: 1) planned error: the positional distance between $\bm T_{\text{o}, \text{d}}$ and planned $\bm T_{\text{o}, T}$ from the initial full trajectory planning (not replanning), and 2) execution error: the distance to the goal after the actual execution.
The hyper-parameters, performances in the competition, and additional results and analysis are provided in the \href{https://rgmc-xl-team.github.io/ingrasp_manipulation}{\textbf{Appendix}}. 
Moreover, although beyond the scope of the competition, our approach can also handle goal object orientations by assigning appropriate orientation weights in $\bm W_{\rm o}$.

% The hyper-parameters we used in the competition are set as $N_{\rm replan}=8$, $\bm W_{\rm o} = \text{diag}(10, 10, 10, 0.01, 0.01, 0.0)$, and $\bm W_{\rm f} = \text{diag}(10, 10, 10, 0.001, 0.001, 0.001)$;
% for the first trajectory optimization for each waypoint, we set $T=3$ and $\lambda = 4e-4$; 
% for the re-planning, we set $T=1$ and $\lambda = 5e-3$.
% All computations are run on a laptop with an Intel i7-9750H CPU (2.6 Hz) and a 16-GB RAM. 

\subsection{Trajectory Optimization}

\begin{figure} [tb]
  \centering 
    \includegraphics[width=\linewidth]{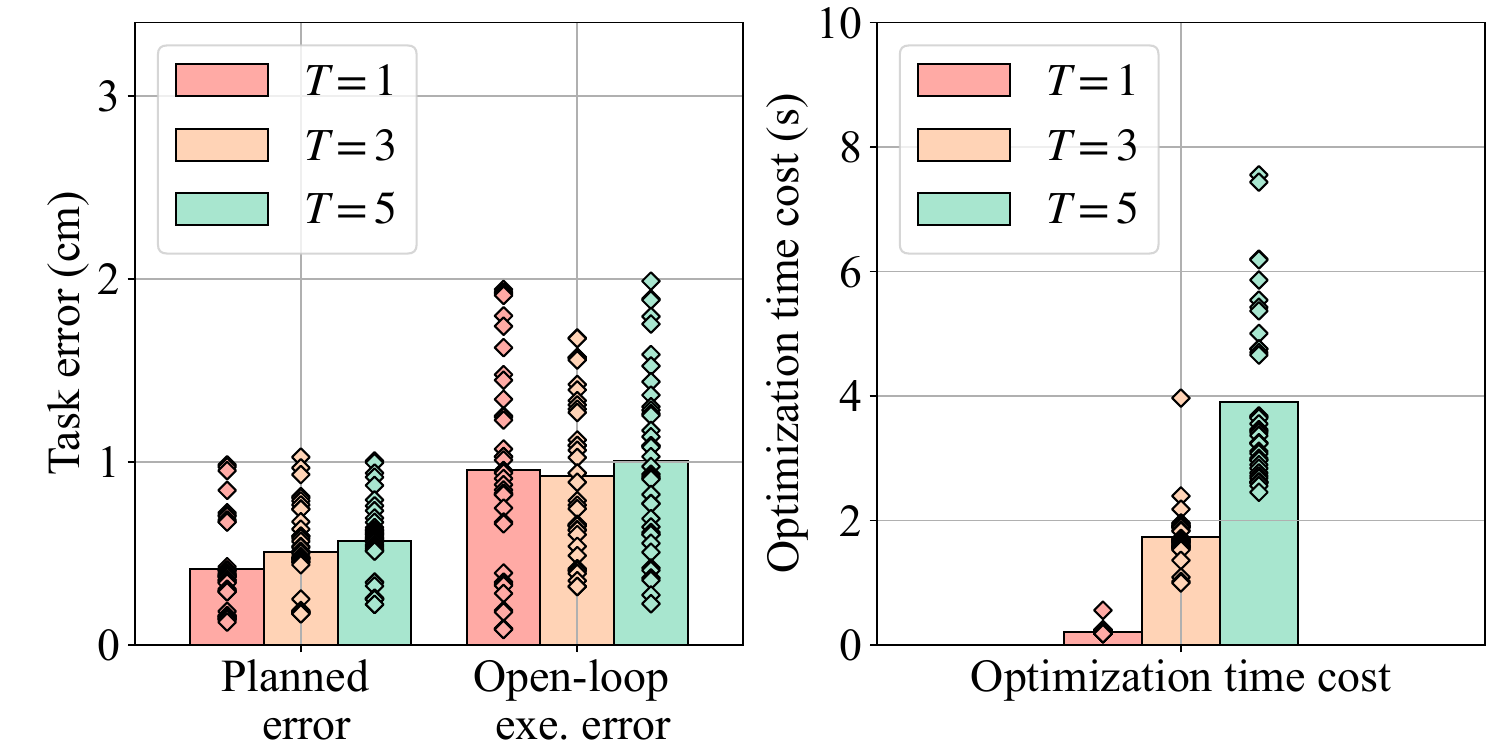} 
  % \vspace{-2mm}
  \caption{Evaluation of the trajectory optimization with different numbers of trajectory steps $T$. Each bar shows the average error/time over 40 waypoints at the corners of the $5 \times 5 \times 5$ (cm) cubic space, and the values for each waypoint are also plotted as the scattered diamond-shaped points.}
  \label{fig:traj_length}
    % \vspace{-5mm}
\end{figure}

We first validate the trajectory optimization itself. 
We choose the eight corners of the $5 \times 5 \times 5$ (cm) cubic space as the most representative and challenging goal object positions, and we task the hand to manipulate the cylinder to continuously reach these eight corners for five iterations (giving a total of 40 waypoints without human intervention).
No replanning is involved. 
The planned error, open-loop execution error, and optimization time cost are shown in Fig. \ref{fig:traj_length}, where we also explore the effects of the number of trajectory steps $T$.

The results show that the actual execution errors of the terminal object pose are larger than the planned errors, primarily due to the simplification of rolling in the trajectory optimization and other practical factors; but the open-loop execution errors remain acceptable, averaging less than 1 cm across the 40 waypoints. 
Second, the different choices of the trajectory steps $T$ have little impact on the execution error. This may be attributed to the physical compliance from the low-level PD controller and the soft fingers, which increase the tolerance of fingertip positions along the trajectory. 
However, when $T=1$, we observe that the fingers may excessively compress the object, as the optimization does not account for in-trajectory grasping.
Third, the time cost of optimization increases with the number of trajectory steps $T$.
As a trade-off between resolution and efficiency, we choose $T=3$ for the competition and subsequent experiments. 
On average, it takes approximately 2 s to plan a trajectory on a laptop with an Intel i7-9750H CPU (2.6 GHz) and a 16-GB RAM.

\textit{Remark}: The variance in task errors of the same approach can be attributed to factors such as different cubic corners, different iterations, and slight differences between initial grasps, which are experimentally investigated in the \href{https://rgmc-xl-team.github.io/ingrasp_manipulation}{\textbf{Appendix}}.

\subsection{Closed-Loop Execution} \label{sec:exp_closed_loop}

\begin{figure} [tb]
  \centering 
    \includegraphics[width=\linewidth]{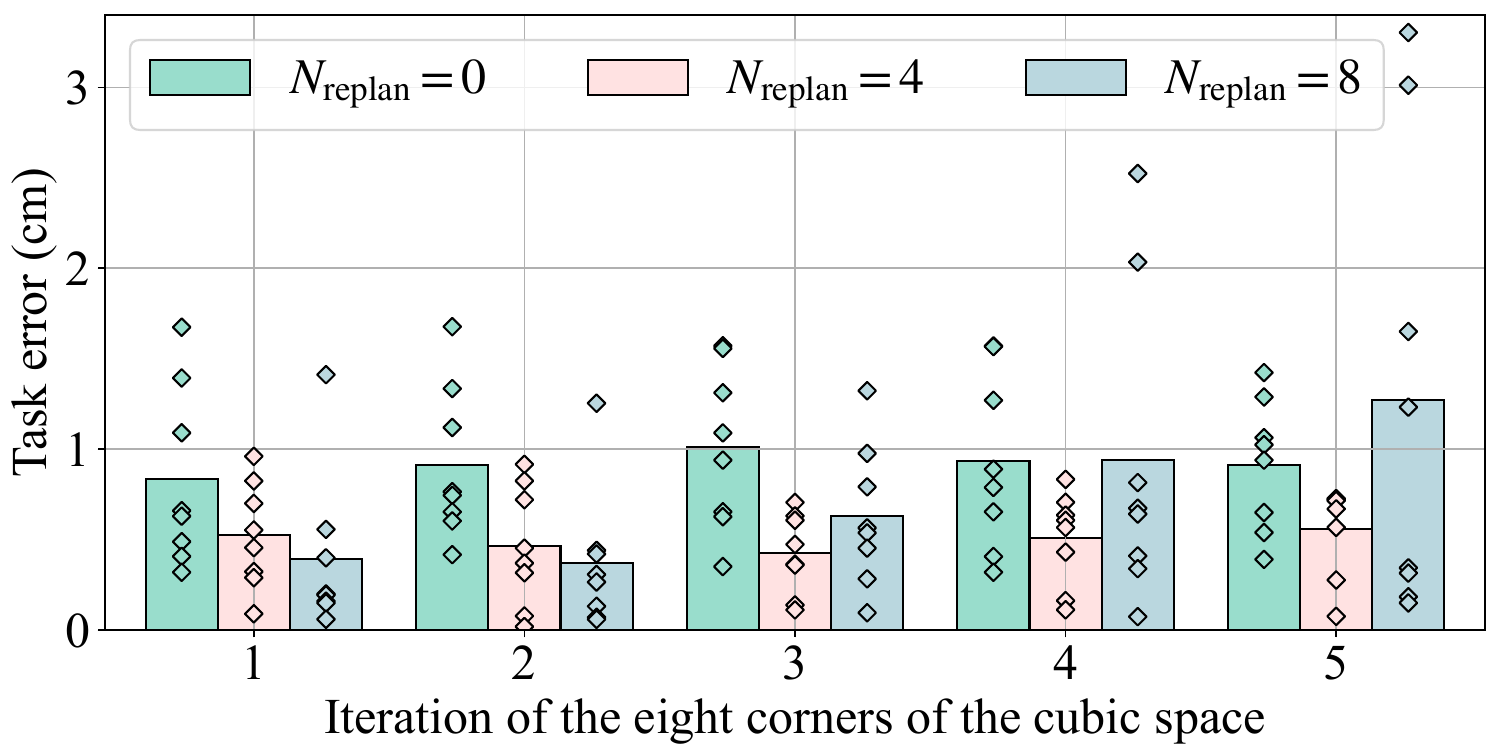} 
  % \vspace{-2mm}
  \caption{Evaluation of the closed-loop execution scheme with different maximum re-planning times $N_{\rm replan}$. Each iteration contains reaching the eight corners of the $5 \times 5 \times 5$ (cm) cubic space, with no human resets between iterations. Each bar shows the average error across the eight corners, and the error for each corner is also plotted as the scattered diamond-shaped points.}
  \label{fig:control_iter}
    % \vspace{-5mm}
\end{figure}

We evaluate the improvements in task accuracy achieved using the closed-loop manipulation scheme. 
We task the hand to manipulate the cylinder to continuously reach the eight corners of the $5 \times 5 \times 5$ (cm) cubic space for five iterations without human intervention. We also test the effects of the maximum re-planning times allowed $N_{\rm replan}$. 

The results in Fig. \ref{fig:control_iter} indicate that: 1) the closed-loop scheme effectively reduces the final object position errors (from approximately 10 mm to approximately 5 mm); 
2) when $N_{\rm replan} \leq 4$, the system continuously reaches the 40 waypoints without a significant drop in accuracy, demonstrating great long-term robustness; and 3) when $N_{\rm replan} = 8$, the accuracy in the first and second iteration exceeds that at lower $N_{\rm replan}$, but the errors in subsequent iterations increase due to slippage between the object and fingertips. 
% Our observations suggest that more times of re-planning increases the risk of reaching ill-conditioned hand-object configurations (e.g., hand singularities or non-tip finger-object contacts). 
Our observations suggest that increased times of re-planning may reduce the contact quality (e.g., slippage or non-tip finger-object contacts). 
Consequently, as a strategy for the competition, we set $N_{\rm replan} = 4$ in the first run to ensure more conservative results and $N_{\rm replan} = 8$ in the second run to aim for higher precision.

\subsection{Comparison with Existing Approach} \label{sec:compare_with_exist}

\begin{figure} [tb]
  \centering 
    \includegraphics[width=\linewidth]{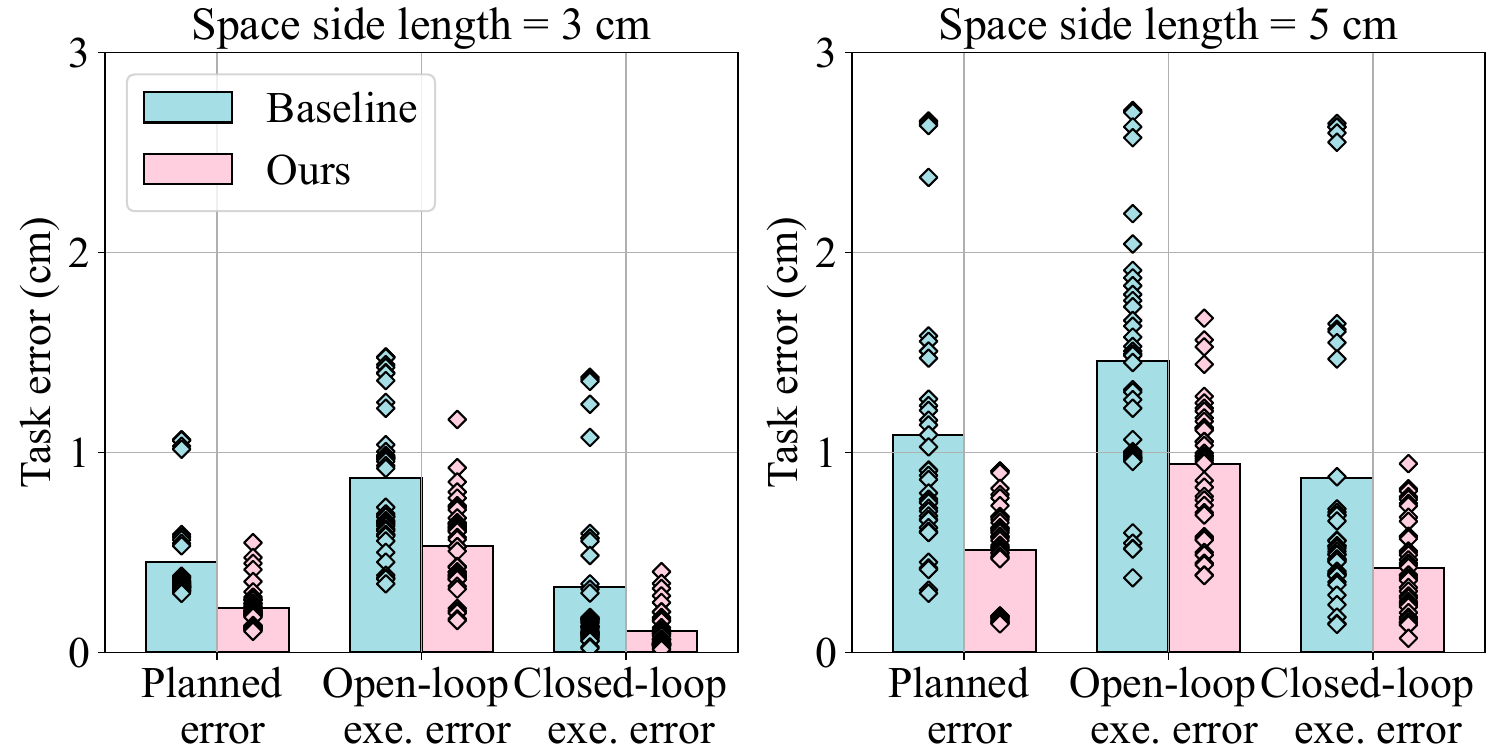} 
  \vspace{-3mm}
  \caption{Comparison between the proposed approach and the baseline \cite{sundaralingam2017relaxed}, which assumes rigid object-thumb contact, whereas we allow for rolling. Each bar shows the average error across 40 waypoints at the corners of the cubic space with side lengths of 3 or 5 cm, and the values of each waypoint are also plotted as the scattered diamond-shaped points.}
  \label{fig:compare_with_baseline}
    \vspace{-3mm}
\end{figure}

Given our task setup involving full-actuated hands, unknown object trajectories, and novel objects, we implement a baseline similar to \cite{sundaralingam2017relaxed} for comparison, which performed the best in the benchmark from \cite{cruciani2020benchmarking}.
% This approach also uses geometry-free kinematic trajectory optimization but assumes rigid contact between the object and thumb-tip. 
Regarding the specific implementation, we adopt the same framework as our approach, whereas the optimization variables include only joint angles, and the object pose is derived from the thumb-tip pose under the rigid contact assumption.
We compare the planned errors, open-loop execution errors, and closed-loop execution errors in continuous five-iteration reaching of the eight corners of both $3 \times 3 \times 3$ and $5 \times 5 \times 5$ (cm) cubic spaces, using the cylinder object. 
We set $N_{\rm replan} = 8$ for the space with a $3$-cm side length, and $N_{\rm replan} = 4$ for the space with a $5$-cm side length.

The results are summarized in Fig. \ref{fig:compare_with_baseline}.
It can be seen that:
1) the planned error is smaller in our approach than in the baseline, since the assumption of rigid object-thumb contact in the baseline limits the theoretical reachable space of the object, whereas our formulation allows rolling contacts;
2) both the open-loop and closed-loop execution errors in our approach are smaller than those in the baseline, demonstrating that our approach also improves the actual execution accuracy in real-world scenarios;
% 3) the gap between the planned and open-loop execution error is smaller in the baseline, as our simplification of rolling contacts for all fingertips introduces a larger sim-to-real gap;
and 3) in the baseline, the execution accuracy at certain corners of the space (e.g., the waypoints with the largest errors) is not significantly improved by closed-loop execution, suggesting that the task errors for these movement directions are primarily due to the limited theoretical reachable space rather than the sim-to-real gap.
% and 4) the accuracy is dramatically improved by closed-loop execution in our formulation, but not in the baseline, suggesting that the task errors in the baseline are primarily due to the limited theoretical reachable space rather than the sim-to-real gap.
These performance results motivated us to develop our proposed approach for the competition, instead of relying on the baseline.

\subsection{Reachable Space and Accuracy} \label{sec:reachable_space}

\begin{figure} [tb]
  \centering 
    \includegraphics[width=\linewidth]{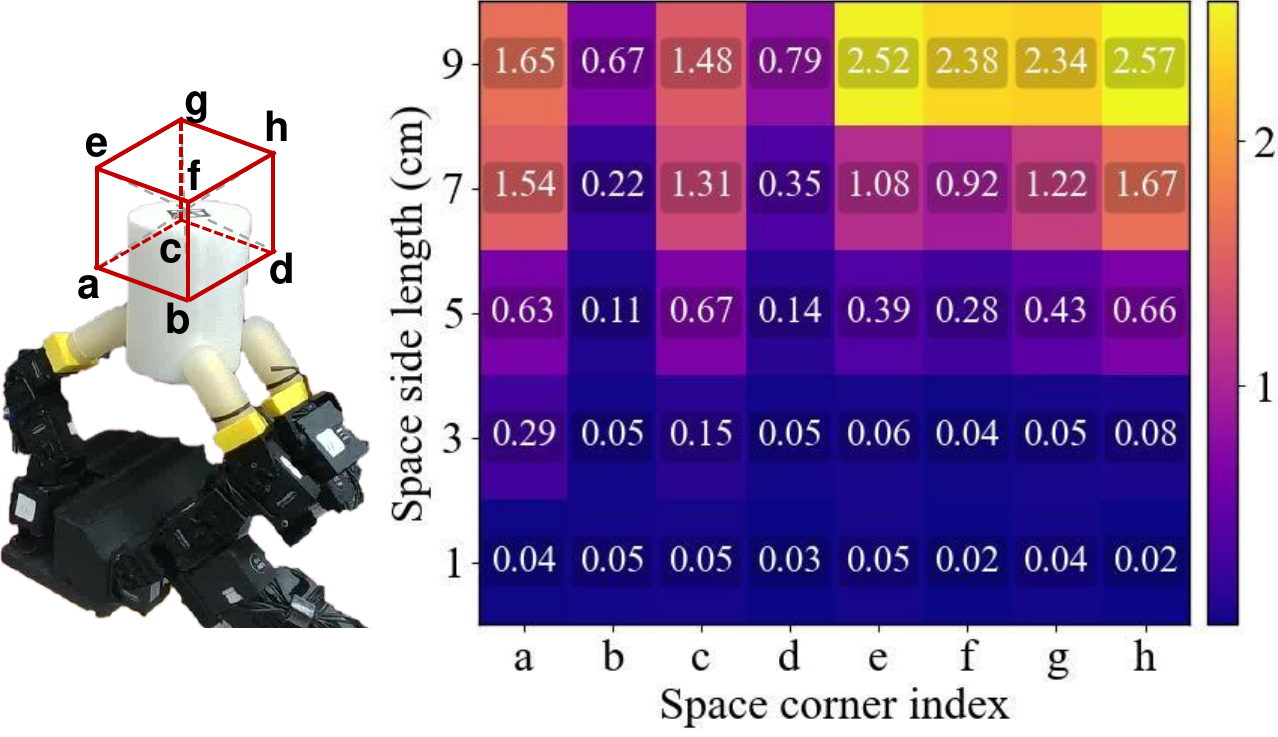} 
  \vspace{-6mm}
  \caption{Evaluation of the object's reachable space and the relationship between goal positions and task accuracy. The left figure illustrates the cubic movement space for the object, with indexed corners and centered at the initial object position. 
  For reference, the side length of the plotted cubic is approximately 6 cm.
  The right figure summarizes the closed-loop execution error for each corner of spaces with side lengths ranging from 1 to 9 cm. The error is averaged over five trials for each goal.}
  \label{fig:reachable_space}
    % \vspace{-5mm}
\end{figure}

We analyze the relationship between the distances to goals and the task accuracy, and further explore the maximum object reachable space. 
The cylinder object is manipulated to reach the corners of cubic spaces with side lengths ranging from 1 to 9 cm. 
The task becomes increasingly challenging as the distance to the goal increases.
Each corner of each cubic space is reached five times. 
For the cubic spaces with side lengths of 7 and 9 cm, we manually adjust the grasp when necessary between goals, as slippage may occur during such large-range movements, potentially affecting subsequent manipulation. 
We manually choose appropriate values of $N_{\rm replan}$ for different goals to achieve the best performance.

The average task error of each goal position and the spatial relationship between the goals and hand is presented in Fig. \ref{fig:reachable_space}.
The results indicate that: 
1) the task accuracy for local goals is very high, with the average error in reaching each corner of the $1 \times 1 \times 1$ (cm) space being no larger than 0.5 mm;
2) the task error increases with the distance to the goal;
3) movements in certain directions are more challenging; e.g., in the $5 \times 5 \times 5$ (cm) space, the average task errors for Corners a, c, and h are larger than those for other corners, due to the asymmetrical finger layout and the distinct mechanical configurations of the thumb and other fingers;
and 4) our approach can reach the boundary of a $9 \times 9 \times 9$ (cm) space with an average error of approximately 2 cm, demonstrating its ability to fully exploit the dexterity and workspace of the fingers to achieve goals as accurately as possible within a large in-hand space.

\subsection{Novel Objects}

\begin{figure} [tb]
  \centering 
    \includegraphics[width=\linewidth]{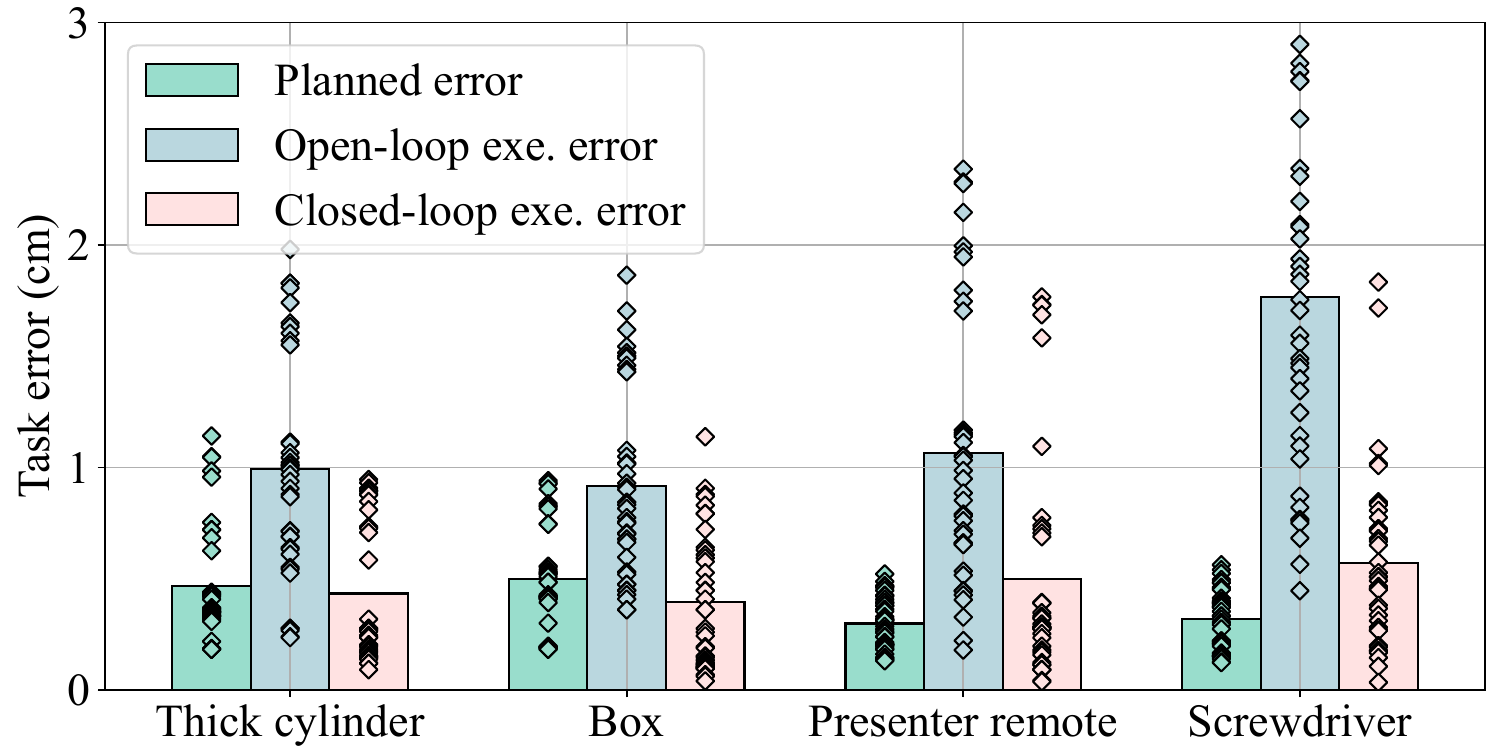} 
  % \vspace{-7mm}
  \caption{Evaluation of the generalization of our approach to novel everyday objects. Each bar shows the average error over 40 waypoints on the corners of the $5\times5\times5$ (cm) space, and the values of each waypoint are also plotted by the scattered diamond-shaped points.}
  \label{fig:novel_objects}
    \vspace{-3mm}
\end{figure}

We further evaluate the generalization of our approach to novel everyday objects, including a thick cylinder lid, box, presenter remote, and screwdriver, as shown in Fig. \ref{fig:objects}(b)-(e). 
We task the hand to manipulate each object to continuously reach the eight corners of the $5 \times 5 \times 5$ (cm) space over five iterations. 
Only for the screwdriver do we occasionally manually adjust the grasp between different goals, due to its thin structure and highly curved surface.

The results in Fig. \ref{fig:novel_objects} indicate that
1) the proposed approach generalizes well to novel everyday objects, with the average task error for each object being approximately 5 mm, even for the challenging screwdriver;
and 2) the curvature of the object surface affects the task accuracy, as high curvature increases the effect of fingertip rolling on contact situations, potentially leading to errors or even unstable grasps.
From our experiments, we conclude that objects with lower weights and curvatures are easier to manipulate using our approach, whereas objects with higher weights and curvatures present greater challenges. This is because increased weight tends to cause slippage, and high curvature causes variations in the grasp contact during rolling.

\section{Conclusion}

This letter proposes a simple and practical approach for in-grasp object movement via trajectory optimization, which won the ICRA 2024 RGMC in-hand manipulation competition and the Most Elegant Solution award. 
Our approach is concise and easy to implement, as it requires no pre-training or object geometries.
Compared with the baseline approach, our approach enlarges the object's reachable space by allowing fingertip rolling.
The quantitative experimental results demonstrate that our approach performs effectively and robustly in the real world, as it can continuously reach 40 waypoints in a large $5 \times 5 \times 5$ (cm) in-hand space with an average error of 5 mm. 
Furthermore, it generalizes well to various everyday objects.
We hope that this letter provides a comprehensive account of our solution and insights gained from the competition, contributing to future research on dexterous in-hand manipulation with a focus on real-world robustness and practicality.

\section{Acknowledgment}
We sincerely thank Professor Kaiyu Hang, Professor Yu Sun, and all organizers of the RGMC for their outstanding support and efforts in making the event a success. Their guidance and dedication were invaluable throughout the competition.
 
\bibliographystyle{IEEEtran}
\bibliography{IEEEabrv, ref}

% \vfill

% \clearpage
% \newpage

%% file: contents/appendix.tex
{\appendices

\section{Analytical Gradients of the Optimization}

In this section, we introduce the key analytical gradients of the trajectory optimization problem.

\subsection{Preliminaries}

According to the theory of Lie groups and Lie algebra for robotics \cite{murray1994mathematical}, we denote the conversion from the axis-angle rotation vector $\bm r \in \mathfrak{so}(3)$ to the rotation matrix $\bm R \in \text{SO}(3)$ as $\bm R = \exp (\bm r^{\wedge})$, and the inverse conversion is denoted as $\bm r = \ln (\bm R)^{\vee}$.

According to the linear approximation of the Baker-Campbell-Hausdorff (BCH) formula \cite{barfoot2024state}, we have
\begin{equation} \label{eq:BCH}
    \ln(\exp(\bm r_1^\wedge)\exp(\bm r_2^\wedge))^\vee \approx
    \left\{\begin{array}{rl}
        \bm J_l(\bm r_2)^{-1} \bm r_1 + \bm r_2, & \hspace{-2mm} \text{when} \, \bm r_1 \rightarrow \bm 0, 
        \\
        \bm J_r(\bm r_1)^{-1} \bm r_2 + \bm r_1, & \hspace{-2mm} \text{when} \, \bm r_2 \rightarrow \bm 0.
    \end{array}\right.
\end{equation}
Here, $\bm J_r(\bm r) = \bm J_l(- \bm r)$, and $\bm J_l(\bm r)$ can be calculated as 
\begin{equation} \label{eq:J_l}
\bm J_l(\bm r) = \frac{\sin \theta}{\theta} \bm I + \left(1-\frac{\sin \theta}{\theta} \right) \bm a \bm a^\transpose + \frac{1-\cos \theta}{\theta} \bm a^{\wedge},
\end{equation}
\begin{equation} \label{eq:J_l_inv}
\bm J_l(\bm r)^{-1} 
= \frac{\theta}{2} \cot \frac{\theta}{2} \bm I 
+ \left(1 - \frac{\theta}{2} \cot \frac{\theta}{2} \right) \bm a \bm a^\transpose 
- \frac{\theta}{2} \bm a^{\wedge},
\end{equation}
where $\theta$ and $\bm a$ are the angle and axis of $\bm r$, respectively.

\subsection{Gradients of Orientation Distances}

Here we generally introduce the gradient regarding the orientation distance. 
We define the weighted scalar distance of orientations $\bm R$ and $\bm R_{\rm d}$ as 
\begin{equation}
d_{\rm r}(\bm R, \bm R_{\rm d}, \bm W_{\rm r}) = \frac{1}{2} \bm r_{\rm e}^\transpose \bm W_{\rm r} \bm r_{\rm e} ,
\end{equation}
where 
$\bm r_{\rm e}$ is defined as $\ln \left( \bm R \bm R_{\rm d}^{-1} \right)^{\vee}$. Note that $\bm r_{\rm e}$ is defined in the world frame $\mathcal{W}$. In addition, $\bm W_{\rm r}$ is a semi-positive definite matrix for weighting. 
Here, $\bm R$ is determined by a variable $\bm x$, and $\bm R_{\rm d}$ is a constant desired orientation.
Note that $\bm r_{\rm e}$ is defined in the same frame as $\bm R$ and $\bm R_{\rm d}$. 

In the following derivation, for convenience, we use both $\bm R \in \text{SO}(3)$ and $\bm r \in \mathfrak{so}(3)$ to represent the same rotation; e.g., $\bm R = \exp (\bm r^{\wedge})$ and $\bm R_{\rm d} = \exp (\bm r_{\rm d}^{\wedge})$.

The gradient of $d_{\rm r}$ w.r.t. the variable $\bm x$ is derived as
\begin{equation} \label{eq:partial_dr_partial_x}
\begin{aligned}
\frac{\partial d_{\rm r}}{\partial \bm x} = 
\frac{\partial d_{\rm r}}{\partial \bm r_{\rm e}} 
\frac{\partial \bm r_{\rm e}}{\partial\bm x} 
= \frac{\partial d_{\rm r}}{\partial \bm r_{\rm e}} 
\frac{\partial \bm r_{\rm e}}{\partial \bm \varphi} 
\frac{\partial \bm \varphi}{\partial \bm x} 
\end{aligned}
\end{equation}
For convenience of the following calculation, here we introduce a perturbation variable $\bm \varphi \in \mathfrak{so}(3)$, which means that we left perturb $\bm R$ by $\Delta \bm R$ (i.e., $(\Delta \bm R)\bm R$ ), where $\Delta \bm R = \exp(\bm \varphi^{\wedge})$.

First, it is easy to obtain
\begin{equation}
    \frac{\partial d_{\rm r}}{\partial \bm r_{\rm e}} 
    =  \bm r_{\rm e}^\transpose \bm W_{\rm r}
\end{equation}

Second, regarding $\frac{\partial \bm r_{\rm e}}{\partial \bm \varphi}$, we have
\begin{equation} \label{eq:partial_re_partial_phi}
\begin{aligned}
    \frac{\partial \bm r_{\rm e}}{\partial \bm \varphi} 
    & = \lim_{\bm \varphi \rightarrow \bm 0}
    \frac{\ln \left( \exp(\bm \varphi^{\wedge}) \exp(\bm r^{\wedge}) \left(\exp (\bm r_{\rm d}^{\wedge}) \right)^{-1} \right)^{\vee}}
    { \bm \varphi}
    \\
    & \quad\quad\quad - \frac{\ln \left( \exp(\bm r^{\wedge}) \left(\exp (\bm r_{\rm d}^{\wedge}) \right)^{-1} \right)^{\vee}}
    { \bm \varphi}
    \\
    & =  \lim_{\bm \varphi \rightarrow \bm 0}
    \frac{ \ln \left( \exp(\bm \varphi^{\wedge}) \exp(\bm r_{\rm e}^{\wedge}) \right)^{\vee} - \ln \left( \exp(\bm r_{\rm e}^{\wedge}) \right)^{\vee} }
    {\bm \varphi}
\end{aligned}
\end{equation}
It follows from the BCH formula (\ref{eq:BCH}) that
\begin{equation} \label{eq:partial_re_partial_phi_2}
\begin{aligned}
    \frac{\partial \bm r_{\rm e}}{\partial \bm \varphi} 
    & =  \lim_{\bm \varphi \rightarrow \bm 0}
    \frac{ \bm J_l(\bm r_{\rm e})^{-1} \bm \varphi +  \bm r_{\rm e} - \bm r_{\rm e} }
    {\bm \varphi}
    = \bm J_l(\bm r_{\rm e})^{-1}
\end{aligned}
\end{equation}
We thus obtain 
$\frac{\partial d_{\rm r}}{\partial \bm \varphi} = \bm r_{\rm e}^\transpose \bm W_{\rm r} \bm J_l(\bm r_{\rm e})^{-1}$. 
Moreover, using (\ref{eq:J_l_inv}), we can obtain $\bm r_{\rm e}^\transpose \bm J_l(\bm r_{\rm e})^{-1} = \bm r_{\rm e}^\transpose$. 
Thus, in special cases where the weights for each orientation dimension are the same (i.e., $\bm W_{\rm r} = w \bm I$), we further have $\frac{\partial d_{\rm r}}{\partial \bm \varphi} = w  \bm r_{\rm e}^\transpose \bm J_l(\bm r_{\rm e})^{-1} = w  \bm r_{\rm e}^\transpose$.

Third, as the perturbation variable $\bm \varphi \rightarrow \bm 0$, we have $\frac{\partial \bm \varphi}{\partial \bm x} = \bm J_{\rm a}(\bm x)$,
where $\bm J_{\rm a}(\bm x)$ is the space Jacobian that relates the spatial angular velocity to $\dot{\bm x}$.

\subsection{Gradients of Pose Distances}
Similar to the derivation of the gradients of orientation distances, the general formula of the gradients of pose distances is derived as follows.

The weighted scalar distance between poses $\bm T$ and $\bm T_{\rm d}$ is defined as  
\begin{equation} 
\begin{aligned}
    d(\bm T, \bm T_{\rm d}, \bm W) = \frac{1}{2} \bm e^\transpose \bm W \bm e
\end{aligned}
\end{equation}
where $\bm e = [\bm p_{\rm e}; \bm r_{\rm e}]$, in which  
$\bm p_{\rm e} = \bm p - \bm p_{\rm d}$ and 
$\bm r_{\rm e} = \ln \left( \exp(\bm r^{\wedge}) \left(\exp (\bm r_{\rm d}^{\wedge}) \right)^{-1} \right)^{\vee}$.
Here, $\bm T$ is determined by a variable $\bm x$, and $\bm T_{\rm d}$ is a constant desired pose.
Note that $\bm e$ is defined in the same frame as $\bm T$ and $\bm T_{\rm d}$.

Similar to (\ref{eq:partial_dr_partial_x}), we introduce a perturbation variable $\bm \phi \in \mathfrak{se}(3)$. 
Then, the gradient of $d$ w.r.t. $\bm x$ is derived as
\begin{equation} \label{eq:partial_d_partial_x}
    \frac{\partial d}{\partial \bm x} 
    = \frac{\partial d}{\partial \bm e}
    \frac{\partial \bm e}{\partial \bm \phi} 
    \frac{\partial \bm \phi}{\partial \bm x} 
\end{equation}
where 
$\frac{\partial d}{\partial \bm e} = \bm e^\transpose \bm W$
and
\begin{equation}
    \frac{\partial \bm e}{\partial \bm \phi} = 
    \left[
    \begin{array}{cc}
    \bm I & \bm 0 \\
    \bm 0 & \bm J_l(\bm r_{\rm e})^{-1}
    \end{array}
    \right]
\end{equation}
Additionally, we have $\frac{\partial \bm \phi}{\partial \bm x} = \bm J(\bm x)$, where $\bm J(\bm x)$ is the space Jacobian that relates the spatial twist to $\dot{\bm x}$.

\subsection{Gradients of $\mathcal{J}_{\rm object}$}

It is easy to see that $\mathcal{J}_{\rm object}$ is only relevant to the object pose at time $T$. The gradient between the position distance cost and the object position variable is easy to derive. Here, we introduce the gradient regarding the orientation distance (i.e., $\frac{\partial d_{\rm r}}{\partial \bm r_{\text{o}, T}}$). 
For brevity, we omit the subscripts $\rm o$ and $T$.

Similar to (\ref{eq:partial_re_partial_phi}) and (\ref{eq:partial_re_partial_phi_2}), we derive that 
\begin{equation} \label{eq:partial_r_partial_phi}
    \frac{\partial \bm \varphi}{\partial \bm r}
    =
    \left( \frac{\partial \bm r}{\partial \bm \varphi} \right)^{-1}
    = \bm J_l(\bm r)
\end{equation}
We then have
\begin{equation}
    \frac{\partial d_{\rm r}}{\partial \bm r} 
    = \bm r_{\rm e}^\transpose \bm W_{\rm r} \bm J_l(\bm r_{\rm e})^{-1}
    \frac{\partial \bm \varphi}{\partial \bm r} 
    = \bm r_{\rm e}^\transpose \bm W_{\rm r} \bm J_l(\bm r_{\rm e})^{-1}
    \bm J_l(\bm r)
\end{equation}

\subsection{Gradients of $\mathcal{J}_{\rm finger}$}

We denote the Lie algebra corresponding to the object pose $\bm T_{\text{o}, t} \in SE(3)$ as $\bm \xi_{\text{o}, t} = [\bm p_{\text{o}, t}; \bm r_{\text{o}, t}] \in \mathfrak{se}(3)$, which is defined in $\mathcal{W}$. 
The optimization variable related to $d({^{\mathcal{O}} \bm T_{i, t}}, {^{\mathcal{O}} \bm T_{i, 0}}, \bm W_{\rm f})$ contains the object pose $\bm \xi_{\text{o}, t}$ and finger joint angle $\bm q_{i, t}$.
For brevity, we omit the subscripts $\rm o$, $i$, and $t$.
We further denote $\bm \vartheta = [\bm \xi; \bm q]$.
% Note that the poses used for calculating the distance in (\ref{eq:J_finger}) are expressed in the object frame $\mathcal{O}$, but here we omit the superscript $\mathcal{O}$ for writing simplicity.

We can use (\ref{eq:partial_d_partial_x}) to calculate the gradient, but we still need to know the space Jacobian that relates the fingertip twist in $\mathcal{O}$ to $\dot{\bm \vartheta}$. 
As the object frame $\mathcal{O}$ is moving, this Jacobian is a relative Jacobian between the finger and object. 
We can calculate this relative Jacobian using individual manipulator Jacobians defined in $\mathcal{W}$ \cite{jamisola2015more}, in which we regard the object as a virtual manipulator.
According to (2) in \cite{jamisola2015more}, the relative Jacobian between the fingertip twist in $\mathcal{O}$ and $\dot{\bm \vartheta}$ can be expressed as 
\begin{equation} \label{eq:relative_jacobian}
{^{\mathcal{O}}\bm J_{\rm f}(\bm \vartheta)} =
\left[
\begin{array}{cc}
    - {^{\mathcal{O}}  \bm \Psi_{\rm f}} 
    {^{\mathcal{O}}  \bm \Omega_{\rm w}} 
    \bm J_{\rm o}(\bm \xi)
    & 
    {^{\mathcal{O}}  \bm \Omega_{\rm w}} 
    \bm J_{\rm f}(\bm q)
\end{array}
\right] ,
\end{equation}
where $\bm J_{\rm o}(\bm \xi)$ is the space Jacobian that relates the object's twist in $\mathcal{W}$ to $\dot{\bm \xi}$, and $\bm J_{\rm f}(\bm q)$ is the space Jacobian that relates the fingertip's twist in $\mathcal{W}$ to $\dot{\bm q}$. 
Similar to (\ref{eq:partial_r_partial_phi}), it can be obtained that
\begin{equation}
    \bm J_{\rm o}(\bm \xi) = 
    % \frac{\partial \bm \phi}{\partial \bm \xi} =
    \left[
    \begin{array}{cc}
      \bm I   &  \bm 0 \\
       \bm 0  &  \bm J_l(\bm r)
    \end{array}
    \right] ,
\end{equation}
where $\bm r$ refers to the object orientation in $\mathcal{W}$. The finger Jacobian $\bm J_{\rm f}(\cdot)$ can be obtained from the finger's kinematics.
The transformation matrices $\bm \Psi$ and $\bm \Omega$ are defined as
\begin{equation}
    {^{a}  \bm \Psi_{b}} = 
    \left[
    \begin{array}{cc}
       \bm I  & -\bm S({^a \bm p_b}) \\
        \bm 0 & \bm I 
    \end{array}
    \right]
    , \quad 
    {^{a}\bm \Omega_{b}} = 
    \left[
    \begin{array}{cc}
       {^{a}\bm R_{b}}  & \bm 0 \\
        \bm 0 & {^{a} \bm R_{b}}
    \end{array}
    \right] ,
\end{equation}
where $\bm S(\bm p)$ refers to the skew-symmetric matrix of vector $\bm p$.

\subsection{Other Gradients}
Other gradients, including the gradients of $\mathcal{J}_{\rm joint}$ and those of the constraints, can be easily derived. The details are omitted here for brevity.

\section{Additional Details of the Competition}

\subsection{Hyper-Parameters}

The hyper-parameters we used in the competition were set as $\bm W_{\rm o} = \text{diag}(10, 10, 10, 0.01, 0.01, 0.0)$, and $\bm W_{\rm f} = \text{diag}(10, 10, 10, 0.001, 0.001, 0.001)$;
for the first trajectory optimization for each waypoint, we set $T=3$ and $\lambda = 4e-4$; 
for the re-planning, we set $T=1$ and $\lambda = 5e-3$. 
These parameters are also used for the experimental evaluation in Section V.

As analyzed in Section V-B, for the competition, we set $N_{\rm replan} = 4$ in the first run to ensure more conservative results and $N_{\rm replan} = 8$ in the second run to aim for higher precision.

\subsection{Performance Results}

The performance details of our approach in the RGMC are provided by the competition organizers, which are summarized in the following tables.
Specifically, the positions of the ten goal waypoints in the competition are listed in Table \ref{tab:goal_waypoints_rgmc}, ranging from (-2.5, -2.5, -2.5) to (2.5, 2.5, 2.5) (cm).
The average task error of the ten waypoints in each run is shown in Table \ref{tab:results_rgmc}. Note that the task error of each individual waypoint is not provided separately, as the organizer did not record them in detail. 
From the results, it can be seen that the precision of the second run is higher than that of the first run. This improvement is attributed to the different choices of $N_{\rm replan}$.

\begin{table}
\centering
\caption{Ten goal waypoints in the competition.}
\label{tab:goal_waypoints_rgmc}
\begin{tabular}{cc} 
\toprule
Waypoint index & Position (x, y, z) (cm) \\ 
\hline
1 & (2.5, 2.5, 0) \\
2 & (2.5, 2.5, 2.5) \\
3 & (-2.5, -2.5, -2.5) \\
4 & (-1.3, -2.0, 0.6) \\
5 & (-1.2, 0.7, 0.6) \\
6 & (0.6, 0.4, 0.2) \\
7 & (0.9, -1.2, -1.3) \\
8 & (-2.0, 2.0, 2.0) \\
9 & (0.0, 0.0, 2.0) \\
10 & (0.0, 0.0, 0.0) \\
\bottomrule
\end{tabular}
\end{table}

\begin{table}[!h]
\centering
\caption{Final results in the competition.}
\label{tab:results_rgmc}
\begin{threeparttable}[b]
\begin{tabular}{cc|c} 
\toprule
 &  & Average task error (cm) \\ 
\hline
\multirow{2}{*}{Cylinder Object} & Run 1 & 0.080 \\
 & Run 2 & 0.054 \\ 
\hline
\multirow{2}{*}{Novel Object\tnote{a}} & Run 1 & 0.125 \\
 & Run 2 & 0.063 \\
\bottomrule
\end{tabular}
\begin{tablenotes}
     \item[a] A mustard bottle from the YCB Dataset, as shown in Fig. \ref{fig:hardware})
\end{tablenotes}
\end{threeparttable}
\end{table}

\section{Additional Results and Analysis}

\subsection{Analysis of Task Error Variance}

\begin{figure*} [tb]
  \centering 
  \subfigure[]{ 
    \label{fig:der_vertices_edges}
    \includegraphics[width=0.48\textwidth]{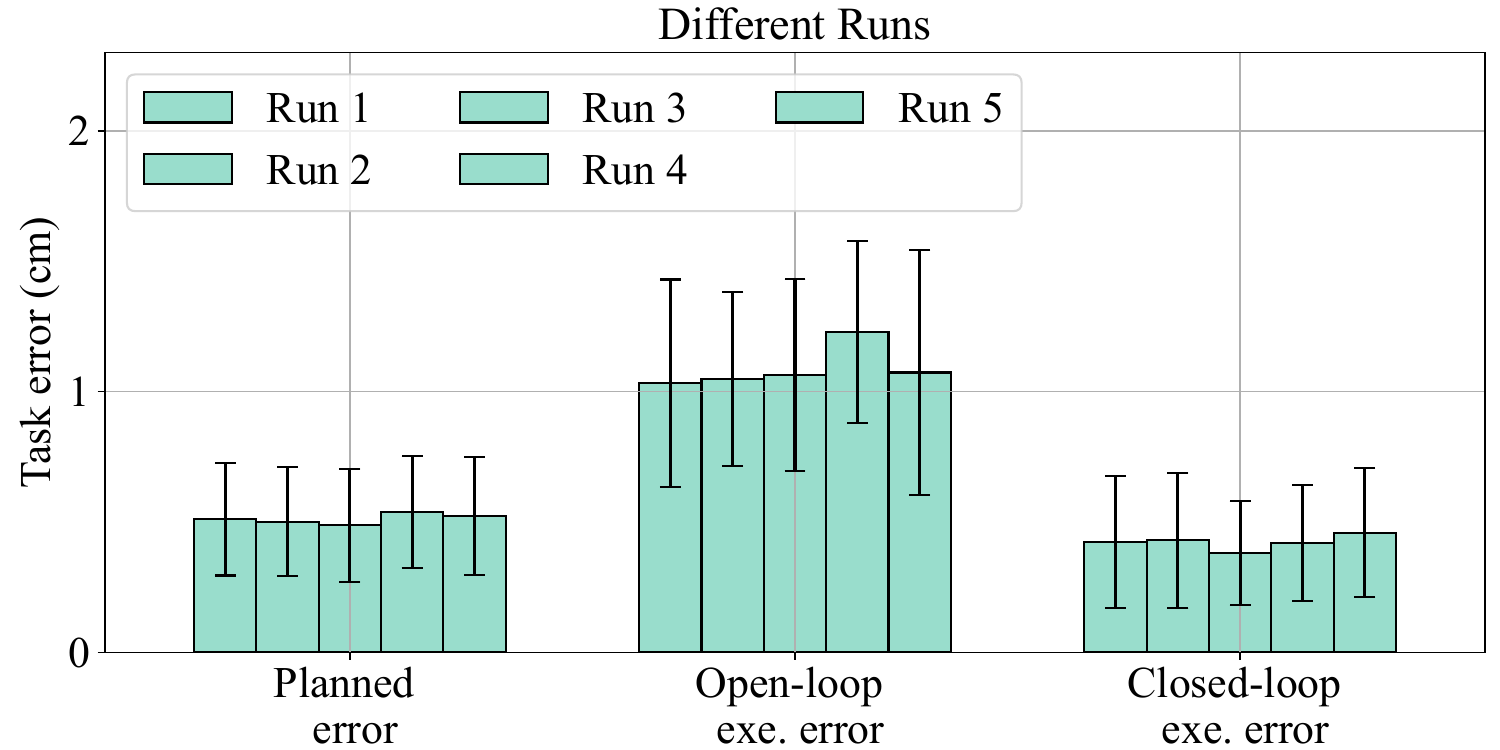} 
  } 
  \subfigure[]{ 
    \label{fig:der_frames}
    \includegraphics[width=0.48\textwidth]{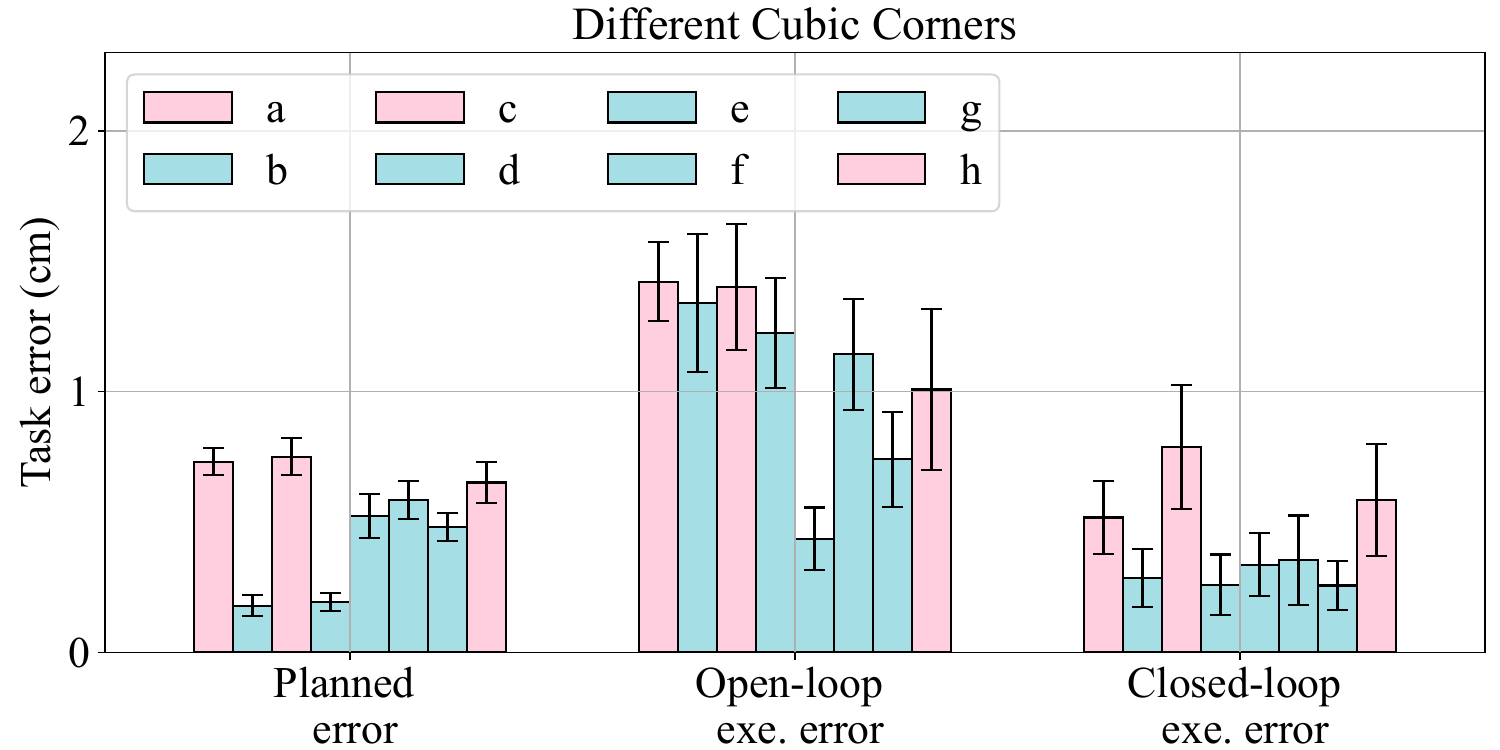} 
  }
  \subfigure[]{ 
    \label{fig:der_frames}
    \includegraphics[width=\textwidth]{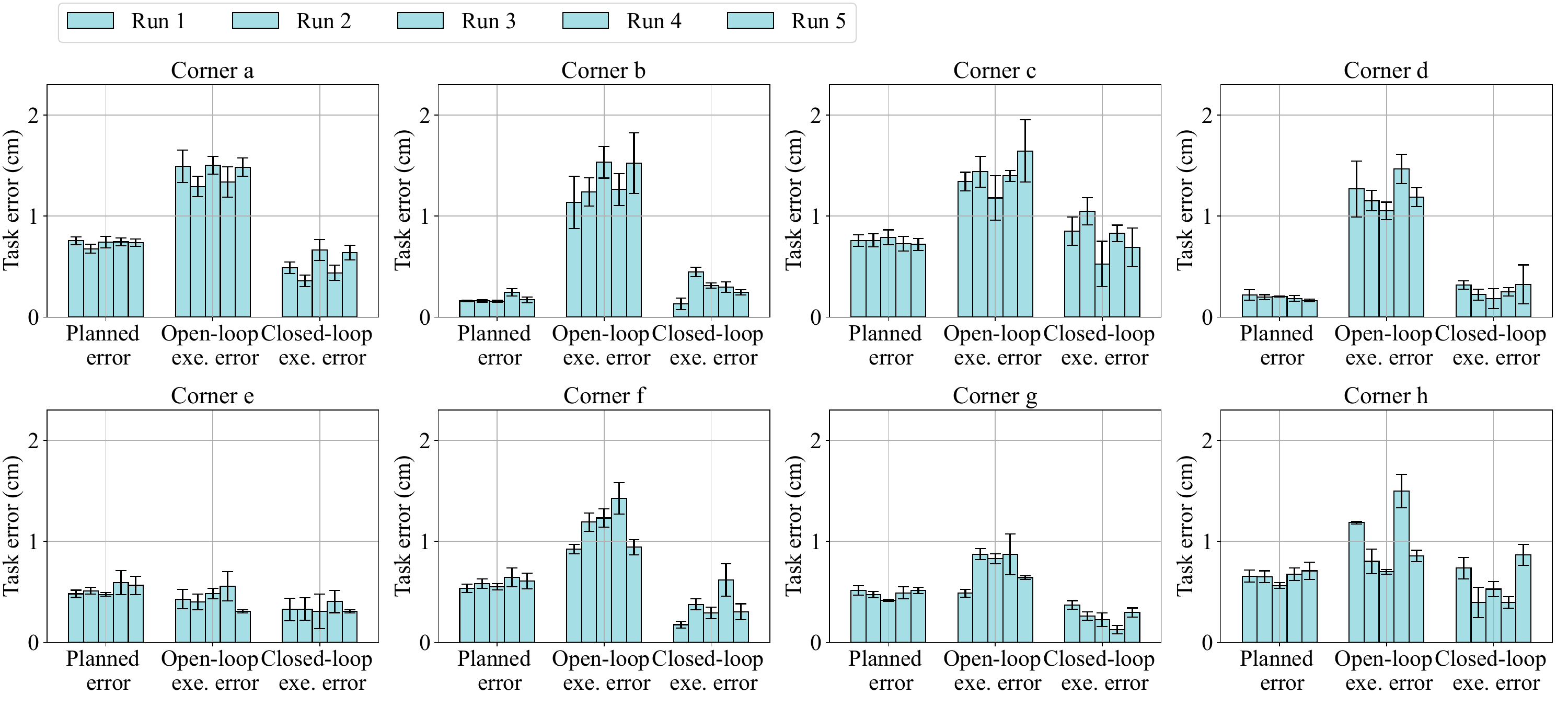} 
  }
%   \vspace{-2mm}
  \caption{Variance of task errors. (a) Variance w.r.t. different runs, where each bar represents the average error over 40 waypoints (five iterations of eight corners) in each run. (b) Variance w.r.t. different cubic corners, where each bar represents the average error over 25 attempts (five iterations in five runs) at each corner. (c) Variance w.r.t. different runs for each corner, where each bar represents the average error over five iterations for each corner in each run. The error bars represent the standard deviation.
  }
  \label{fig:variance}
    % \vspace{-5mm}
\end{figure*}

The diamond points within Figs. 6, 7, 8 and 10 represent the task error of each waypoint along with the summary statistic. 
Overall, the variance in task errors can be attributed to factors such as different cubic corners, different iterations, and slight differences between initial grasps.
We further investigate the variance through extensive experiments. 
Specifically, we apply the proposed approach to manipulate the object to the corners of the $5 \times 5 \times 5$ (cm) cubic space over five iterations without human intervention. This process was repeated five times, with the initial grasp reset by a human before each trial. 
We provide a video of four initial grasps at \href{https://rgmc-xl-team.github.io/ingrasp_manipulation/initial_grasp.mp4}{\text{url}}\footnote{
\href{https://rgmc-xl-team.github.io/ingrasp_manipulation/initial_grasp.mp4}{https://rgmc-xl-team.github.io/ingrasp\_manipulation/initial\_grasp.mp4}
}.

Fig. \ref{fig:variance}(a) and (b) summarize the average task errors for different runs and different cubic corners. The results indicate that 
1) the average task errors of different runs are relatively consistent, with the gap between the largest and smallest closed-loop error being less than 0.08 cm, demonstrating that the slight differences in initial grasps have little impact on the overall statistic results;
and 2) certain corners (e.g., a, c, h) exhibit averagely larger errors, as discussed in Section \ref{sec:reachable_space}. 
Then, we further investigate the variance among different runs for each corner, as shown in Fig. \ref{fig:variance}(c). It can be seen that the planned errors for the same corner remains relatively consistent; in contrast, the execution errors for the same corner vary across different runs and iterations. This variability arises from minor differences in the manually established initial grasps and slight changes in the grasp (e.g., slippage) during continuous manipulation.

\subsection{Effect of Moving Back to Initial State}

As described in Section \ref{sec:closed_loop_exe}, we employ a strategy where the fingers return to the initial state (following the forward trajectory) after reaching each waypoint. This approach is adopted because the initial state typically provides a more favorable starting point for trajectory optimization toward the next goal.
To experimentally evaluate the effectiveness of this strategy, we task the hand with manipulating the object to the corners of the $5 \times 5 \times 5$ (cm) cubic space over five iterations. We compare the performance of the approach with and without this strategy, conducting five trials for each condition. 
An example of the manipulation process without returning to the initial state is shown in a video at \href{https://rgmc-xl-team.github.io/ingrasp_manipulation/not_move_back.mp4}{\text{url}}\footnote{
\href{https://rgmc-xl-team.github.io/ingrasp_manipulation/not_move_back.mp4}{https://rgmc-xl-team.github.io/ingrasp\_manipulation/not\_move\_back.mp4}}.

When using this strategy, the average task error of the first iteration (8 corners) is 0.40 cm, and the average task error of all five iteration (40 waypoints) is 0.42 cm. 
When not using this strategy, the average task error of the first iteration is 0.47 cm; however, the object falls in the second iteration in all five tests due to low contact quality. These results indicate that employing this strategy can enhance the robustness in continuous waypoint reaching.

\subsection{Impact of Excessive Re-Planning}

In Section \ref{sec:exp_closed_loop}, we point out that excessive re-planning may degrade contact quality and lead to larger task errors. 
This problem is primarily due to the ``initial state" of the trajectory optimization.
We assume the initial grasp is a stable and manipulable grasp, which holds true for human-designed initial grasps. 
However, during replanning, the ``initial state" of the new optimization problem is set as the terminal state from the previous execution. 
Note that our simple trajectory optimization problem formulation does not explicitly constrain the quality of the finger-object contacts. Instead, it tries to implicitly maintain the contact quality by softly penalizing deviations between the terminal and initial configuration (through the cost term $\mathcal{J}_{\rm finger}$ and $\mathcal{J}_{\rm joint}$).
Consequently, the terminal state from the previous execution may not be as good a grasp as the initial state, leading to a gradual decline in contact quality as the number of replanning iterations increases. 

We provide an example to illustrate this problem in Fig. \ref{fig:infinite_replanning}. It is shown that when applying too many replanning times to reach the goal waypoint, unmodeled contact occurs between the object and non-spherical parts of the fingers, which leads to significant slippage.

\begin{figure} [tb]
  \centering 
    \includegraphics[width=\linewidth]{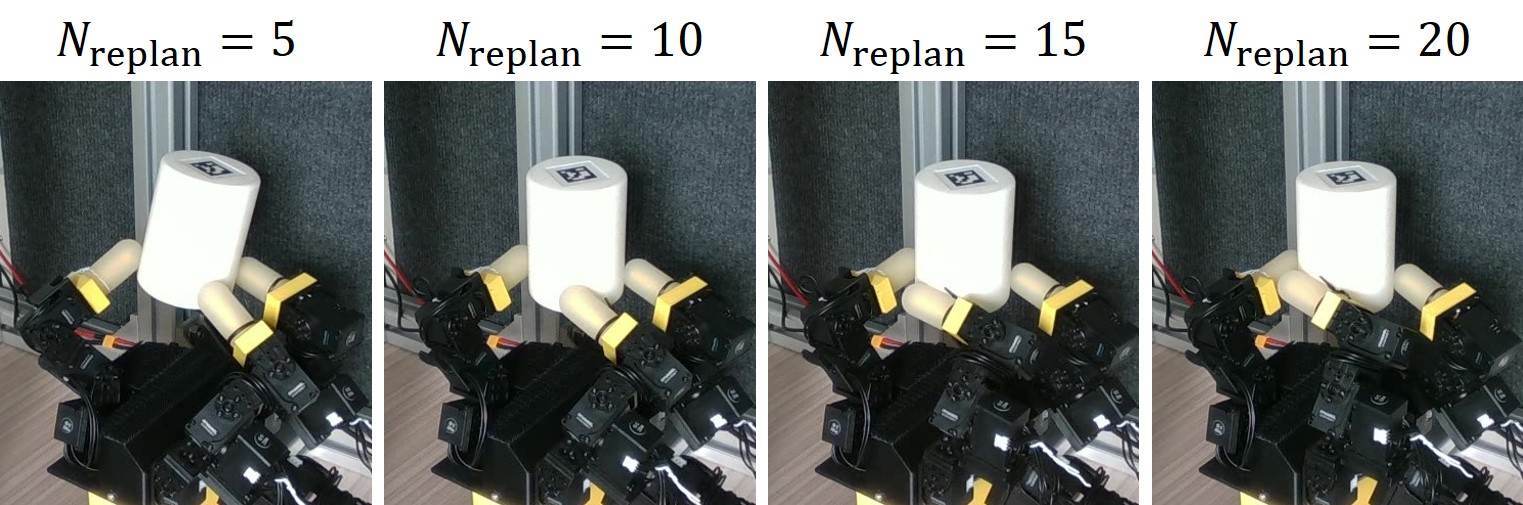} 
  % \vspace{-2mm}
  \caption{Example of applying excessive re-planning.}
  \label{fig:infinite_replanning}
    % \vspace{-5mm}
\end{figure}

\subsection{Effect of Weights for Object Pose Cost}

\begin{figure} [tb]
  \centering 
    \includegraphics[width=\linewidth]{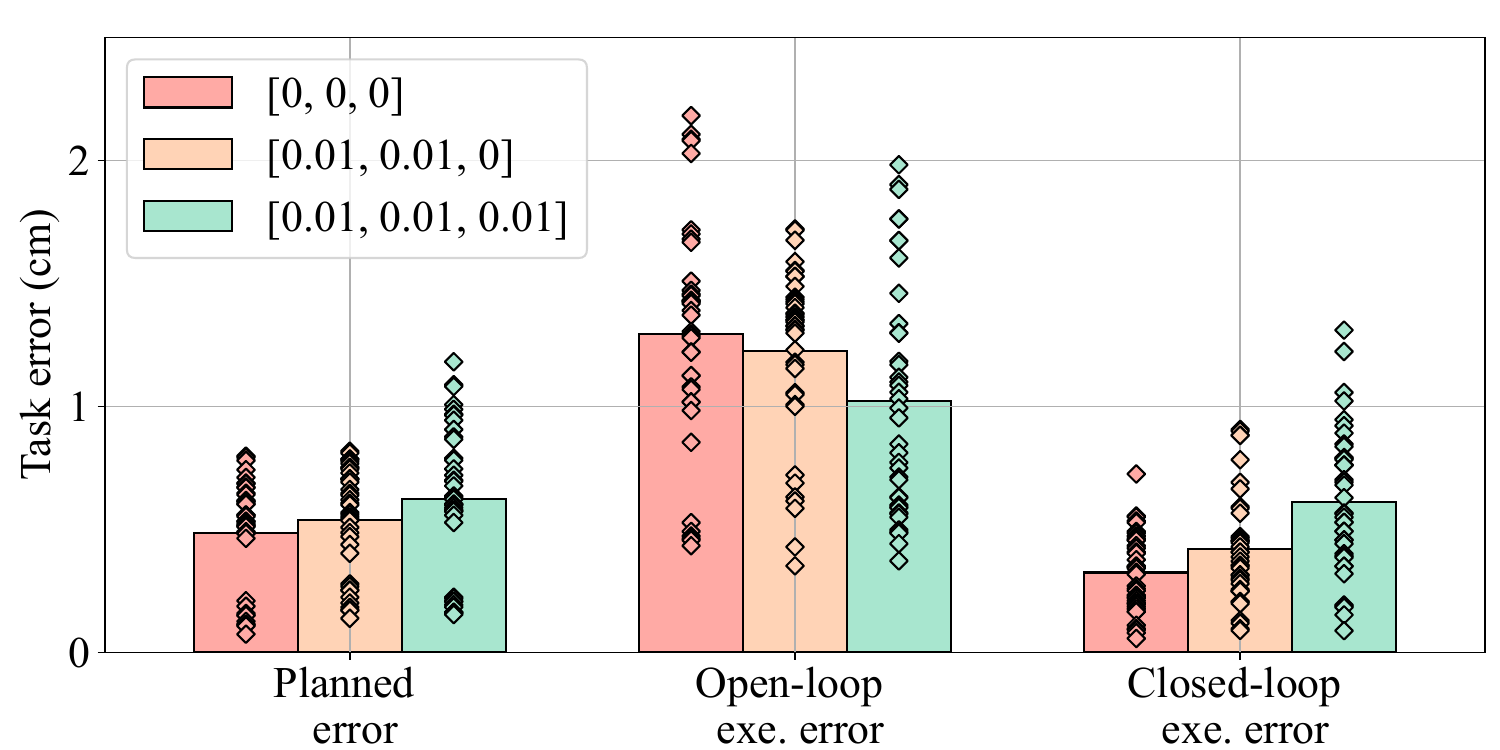} 
  % \vspace{-2mm}
  \caption{Effect of different weights for the object orientation cost. Each bar shows the average error over 40 waypoints on the corners of the $5\times5\times5$ (cm) space, and the values of each waypoint are also plotted by the scattered diamond-shaped points.}
  \label{fig:different_weights}
    % \vspace{-5mm}
\end{figure}

In the competition, we empirically used $\bm W_{\rm o} = \text{diag}(10, 10, 10, 0.01, 0.01, 0.0)$ for the object pose cost $\mathcal{J}_{\rm object}$. We set the goal object orientation as the current (initial) orientation, and applied the non-zero weights to slightly regulate the rotation along the X and Y axes, which improved the manipulation robustness in our experience.

We further quantitatively evaluate the effect of different choices of the weights for the orientation, including $\text{diag}(0.0, 0.0, 0.0)$, $\text{diag}(0.01, 0.01, 0.0)$, and $\text{diag}(0.01, 0.01, 0.01)$. The results of one run (40 waypoints) are shown in Fig. \ref{fig:different_weights}. 
It can be seen that 
1) increasing the regulation on object orientation leads to larger planned and closed-loop errors, as it restricts the object's reachable space; however, it results in smaller open-loop errors, as smaller rotation generally reduces the risk of unexpected slippage during manipulation;
and 2) regulating the Z-axis rotation further increases the closed-loop task errors, compared with regulating only the X and Y axes.
Although the results of this run suggest that no regulation might yield higher precision, we find that significant slippage sometimes occur during continuous manipulation without any regulation. 
Consequently, we chose $\text{diag}(0.01, 0.01, 0.0)$ for the competition as a trade-off between accuracy and robustness.

\subsection{Goals of Object Poses}

We further demonstrate our approach's capability to handle goals of object poses, which include both positions and orientations. The task involves continuous reaching three goal poses, for which we set $\bm W_{\rm o} = \text{diag}(10, 10, 10, 1, 1, 1)$. The details of the goals and closed-loop manipulation results are listed in Table \ref{tab:exp_object_pose}. Snapshots of the manipulation process are shown in Fig. \ref{fig:exp_orientation}. 
The results demonstrate that our approach can be easily adapted to address goal orientations.

\begin{table}
\centering
\caption{Evaluation of our approach with goals of object poses.}
\label{tab:exp_object_pose}
\begin{tabular}{c|cc|cc} \toprule
Index & \begin{tabular}[c]{@{}c@{}}Goal \\translation \\(cm)\end{tabular} & \begin{tabular}[c]{@{}c@{}}Goal \\rotation\\(degree)\end{tabular} & \begin{tabular}[c]{@{}c@{}}Position \\error\\(cm)\end{tabular} & \begin{tabular}[c]{@{}c@{}}Orientation \\error\\(degree)\end{tabular} \\ \hline
1 & (0, -1, -1) & (0, 20, 0) & 0.22 & 1.7 \\
2 & (-2, 0, 0) & (30, 0, 0) & 0.74 & 4.5 \\
3 & (0, 0, 2) & (0, 0, 40) & 0.16 & 1.2 \\ \bottomrule
\end{tabular}
\end{table}

\begin{figure} [tb]
  \centering 
    \includegraphics[width=0.9\linewidth]{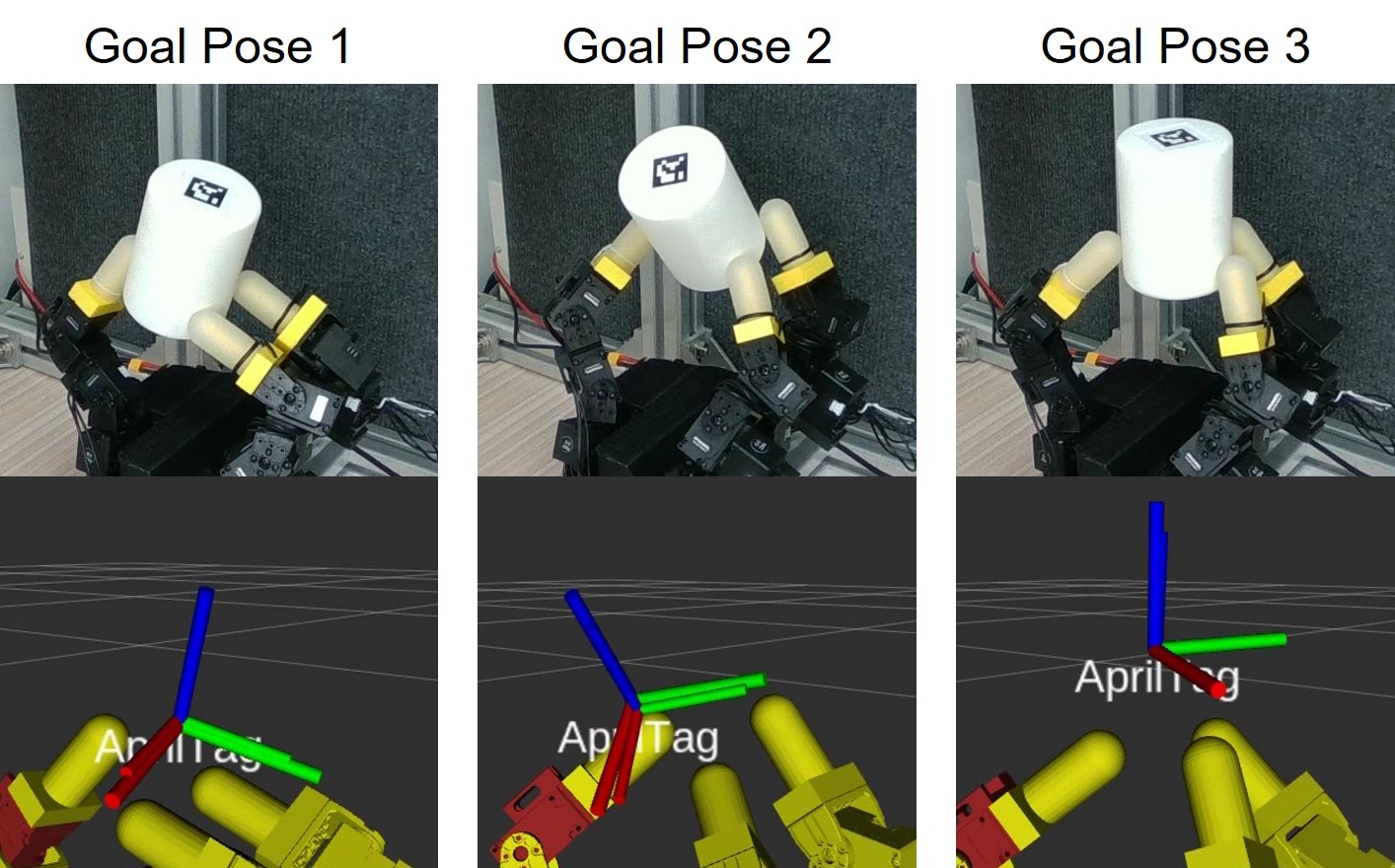} 
  % \vspace{-2mm}
  \caption{Snapshots of reaching goal object poses. The figures in the second row visualize the poses, where the larger axes represent the goal poses and the smaller axes represent the poses of the AprilTag. The manipulation process is also shown in the supplementary video.}
  \label{fig:exp_orientation}
    % \vspace{-5mm}
\end{figure}

\subsection{Implementation of Baseline}

In Section \ref{sec:compare_with_exist}, we implement a baseline similar to \cite{sundaralingam2017relaxed} for comparison. Here we provide the details regarding the re-implementation. 
We adopt the same framework as our approach, whereas the optimization variables include only joint angles, and the object pose is derived from the thumb-tip pose under the rigid contact assumption. 
Our implementation closely follows that in \cite{sundaralingam2017relaxed}, with the following differences: 
1) we do not include the cost of in-trajectory object poses, whose references are obtained by linear interpolation between the start and goal object poses in theirs; 
2) we treat the joint velocity/movement limits as a soft constraint (penalty) instead of a hard constraint in theirs;
and 3) we use the same hyper-parameters as those in our approach.

} % end of appendix